\documentclass[numsec,webpdf,modern,large]{oup-authoring-template}%

\usepackage{subfigure}
\usepackage{wrapfig}
\usepackage{tabularx}
\usepackage{xcolor}

\makeatletter
\renewcommand{\@thesubfigure}{\normalsize(\alph{subfigure})\hskip\subfiglabelskip}
\makeatother

\definecolor{darkgreen}{rgb}{0.0, 0.8, 0.0}

\graphicspath{{Fig/}}

\theoremstyle{thmstyleone}%
\theoremstyle{thmstyletwo}%
\newtheorem{remark}{Remark}%
\theoremstyle{thmstylethree}%

\newcommand{\mysubsection}[1]{\vspace{-8px}\subsection{#1}}

\begin{document}

\journaltitle{Bioinformatics}
\DOI{DOI HERE}
\copyrightyear{2022}
\pubyear{2019}
\access{Advance Access Publication Date: Day Month Year}
\appnotes{Original Paper}

\firstpage{1}


\title[Short Article Title]{Benchmarking drug-drug interaction prediction methods: a perspective of distribution changes
}

\author[1]{Zhenqian Shen}
\author[1]{Mingyang Zhou}
\author[3]{Yongqi Zhang}
\author[1,2,$\ast$]{Quanming Yao\ORCID{0000-0000-0000-0000}}

\authormark{Zhenqian Shen et al.}

\address[1]{\orgdiv{Department of Electrical Engineering}, \orgname{Tsinghua University}}

\address[2]{\orgdiv{State Key laboratory of Space Network and Communications}, \orgname{Tsinghua University}, \orgaddress{ \postcode{100084}, \state{Beijing}, \country{China}}}

\address[3]{\orgdiv{Thrust of Data Science and Analytics}, \orgname{The Hong Kong University of Science and Technology (Guangzhou)}, \orgaddress{ \postcode{511453}, \state{Guangdong}, \country{China}}}

\corresp[$\ast$]{Corresponding author. \href{email:email-id.com}{qyaoaa@tsinghua.edu.cn}}

\received{Date}{0}{Year}
\revised{Date}{0}{Year}
\accepted{Date}{0}{Year}



\abstract{
\textbf{Motivation:} Emerging drug-drug interaction (DDI) prediction is crucial for new drugs 
but is hindered by distribution changes between known and new drugs in real-world scenarios. 
Current evaluation often neglects these changes, relying on unrealistic i.i.d. split due to the absence of drug approval data. \\
\textbf{Results:} 
{ We propose DDI-Ben, a benchmarking framework for emerging DDI prediction under distribution changes. DDI-Ben introduces a distribution change simulation framework that leverages distribution changes between drug sets as a surrogate for real-world distribution changes of DDIs, and is compatible with various drug split strategies. Through extensive benchmarking on ten representative methods, we show that most existing approaches suffer substantial performance degradation under distribution changes. Our analysis further indicates that large language model (LLM) based methods and the integration of drug-related textual information offer promising robustness against such degradation. To support future research, we release the benchmark datasets with simulated distribution changes. Overall, DDI-Ben highlights the importance of explicitly addressing distribution changes and provides a foundation for developing more resilient methods for emerging DDI prediction.} \\
\textbf{Availability and implementation:} Our code and data are available at \url{https://github.com/LARS-research/DDI-Bench}.
}


\maketitle

\section{Introduction}
\label{sec:intro}

With the rapid development of drug discovery, numerous emerging drugs are being developed to treat various diseases.
As these drugs contain novel chemical substances with unknown pharmacological risks, it is crucial to conduct emerging drug-drug interaction~(DDI) prediction to identify not only potential adverse interactions but also opportunities for effective combination therapies.
In clinical experiments, measuring and verifying DDIs are extremely time-consuming and expensive, which motivates recent development of computational methods for the problem of DDI prediction.
Currently, the research on computational methods for DDI prediction problem has been more and more prevalent, and different types of machine learning techniques have been used for that problem, including feature based methods~\citep{rogers2010extended, ryu2018deep, liu2022predict}, embedding based methods~\citep{karim2019drug, yao2022effective}, graph neural network~(GNN) based methods~\citep{zitnik2018modeling, lin2021kgnn, yu2021sumgnn, zhang2023emerging}, Graph-transformer based methods~\citep{su2024dual, chen2024drugdagt}, and large language model~(LLM) based methods~\citep{zhu2023learning, xu2024ddi, abdullahi2025k}.

\begin{figure*}[t]
	\centering
	\vspace{-5px}
	\subfigure[\normalsize Common split.]{
		\includegraphics[width=0.23\textwidth]{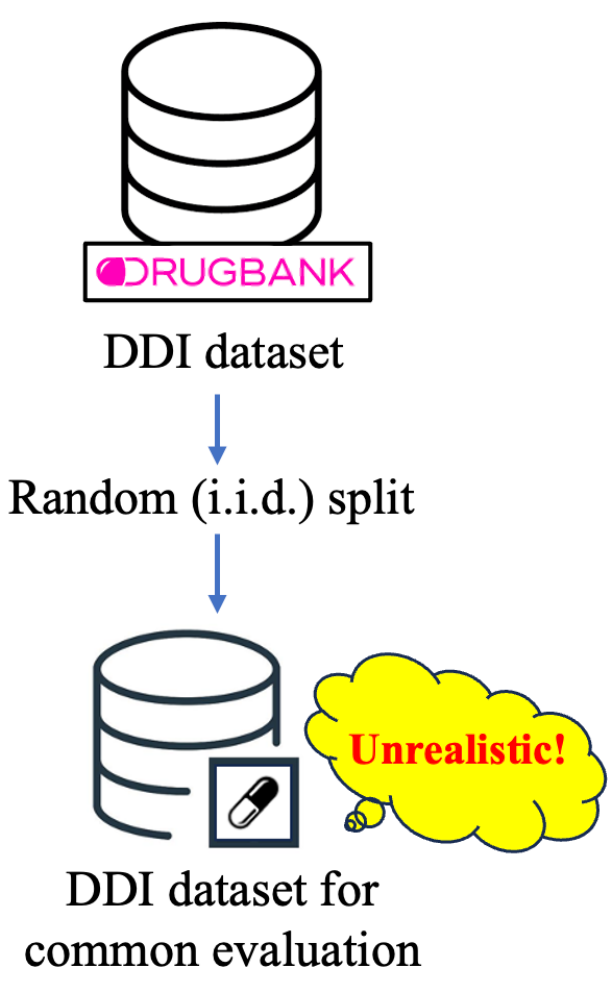}
	}
	\quad
	\subfigure[\normalsize Proposed distribution change simulation framework.]{
		\includegraphics[width=0.55\textwidth]{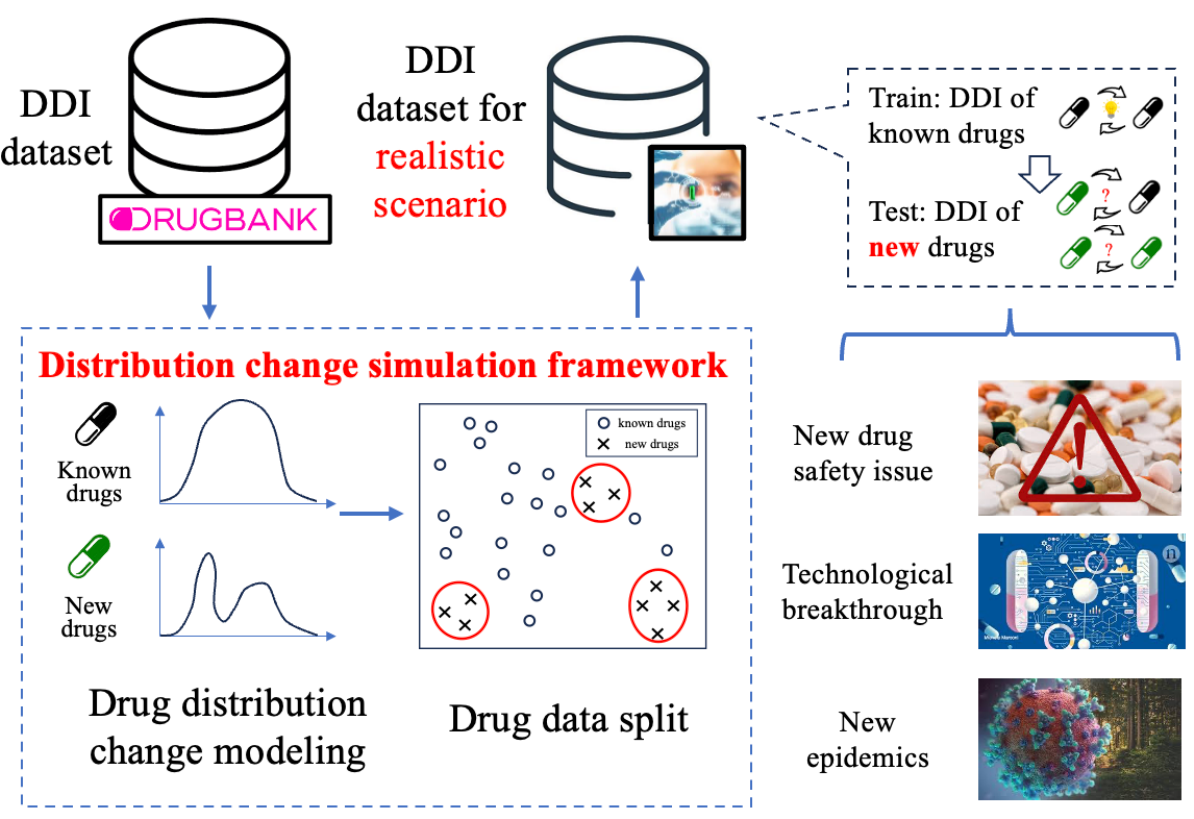}
	}
	\caption{Comparison between common DDI data split and proposed distribution change simulation framework for emerging DDI prediction evaluation.}
	\label{fig:framework_comp}
	\vspace{-15px}
\end{figure*}

In this work, we focus on benchmarking for emerging DDI prediction in a perspective from distribution changes, attempting to reduce the mismatching of existing DDI prediction evaluation and realistic drug development scenarios.
Current evaluation frameworks for emerging DDI prediction methods inadequately address the phenomenon of distribution changes inherent in real-world data, primarily due to following reasons: 
firstly, current widely-used DDI datasets~(e.g. Drugbank~\citep{tatonetti2012data}, 
TWOSIDES~\citep{wishart2018drugbank}) lack information on approval timelines of drugs, making it hard
to directly incorporate drug distribution changes into the evaluation framework.
Secondly, owing to the lack of time information for drugs, most of the existing emerging DDI prediction methods implicitly assume that known drugs and new drugs follow the same distribution~\citep{liu2022predict, zhang2023emerging}, 
dividing drugs into known and new drug sets in an i.i.d. manner~(Figure~\ref{fig:framework_comp}~(a)).
They neglect the phenomenon of distribution changes that is inherent in realistic drug development process, which makes their evaluation results questionable. 
\textbf{In order to evaluate existing DDI prediction methods in a scene closer to real-world scenarios, it is necessary to simulate the distribution changes between known and new drugs.}

{

To fulfill the gap, we propose DDI-Ben, a benchmark for emerging DDI prediction that investigates existing computational methods in a perspective from drug distribution changes. 
We first introduce a distribution change simulation framework~(Figure~\ref{fig:framework_comp}~(b)) that is compatible with different drug split schemes to replicate distribution changes in real-world DDI data. 
We conduct extensive experiments on representative emerging DDI prediction methods, ranging from simple MLP to recent ones based on large language models. 
We observe a significant drop in the performance of existing methods on emerging DDI prediction tasks with distribution change introduced. 
It is also found that large language model, drug-related textual information can be key factors to alleviate the negative impact of that performance drop.
In summary, our main contributions are as follows:

\begin{itemize}
\item DDI-Ben proposes a distribution change simulation framework that uses distribution changes between drug sets as a surrogate to simulate distribution changes in emerging DDI prediction problem. The framework is compatible with various drug split strategies, which can reflect the distribution changes in real-world scenarios. 
\item Through benchmarking evaluations on emerging DDI prediction, DDI-Ben demonstrates that most existing methods lack robustness against distribution changes. Our analysis suggests that developing LLM-based approaches and incorporating drug-related textual information can help mitigate performance degradation under such conditions. 
\item We also release emerging DDI prediction dataset with simulated distribution changes used in our benchmark. 



\end{itemize}
}

	

\section{Emerging DDI Prediction Task Description}
\label{sec:task}

Assume the drug set is $\mathcal{D}$ and the possible interaction set between drugs is $\mathcal{R}$. 
The problem of DDI prediction is to learn a predictor $p : \mathcal{D} \times \mathcal{D} \rightarrow \mathcal{R}$ that can accurately predict the interaction type $r \in \mathcal{R}$ between drugs $(u, v) \in \mathcal{D} \times \mathcal{D}$. 
To conduct emerging DDI prediction evaluation, drug set $\mathcal{D}$ is usually divided into two sets: known drug set $\mathcal{D}_k$ and new drug set $\mathcal{D}_n$. 
We mainly focus on the DDI prediction relevant to emerging~(new) drugs, including two types of tasks.
The S1 task is to determine the DDI type between a known drug and a new drug. 
The S2 task is to determine the DDI type between two new drugs.
As mentioned above, the distribution changes between training and test DDIs are neglected in existing DDI prediction evaluation settings, 
which makes their evaluation results unreliable in realistic drug development scenarios.

\begin{figure}[ht]
    \centering
    \vspace{-12px}
    \includegraphics[width=0.4\textwidth]{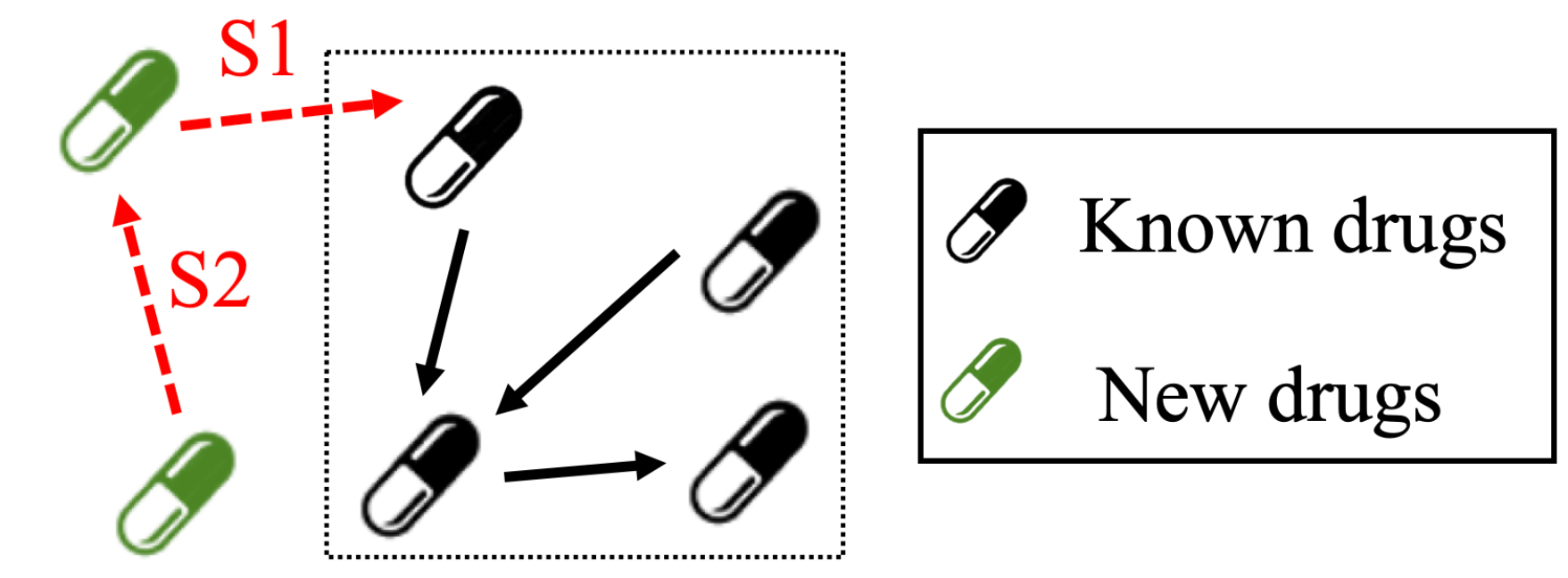}
    \vspace{-8px}
    \caption{Emerging DDI prediction task description.}
    \label{fig:task}
    \vspace{-32px}
\end{figure}

\section{Distribution Change Simulation Framework}
\label{sec:diff_gl}

In emerging DDI prediction, distributional changes between training and test data are critical issues, which primarily stems from distribution changes between known and new drugs~\citep{wishart2018drugbank, celebi2019evaluation}. 
However, existing DDI datasets do not provide approval time information for drugs, making it hard to directly distinguish known and new drugs.
To address this problem, we model distribution changes between known drug set and new drug set as a surrogate to simulate distribution changes. 

In this section, we first introduce a distribution change simulation framework that can reflect realistic scenarios in emerging DDI prediction, with an analysis for understanding of that framework. 
Based on observation of real-world data, we propose cluster-based distribution change modeling for DDI benchmarking evaluation. 
Finally, we provide a comparison among different drug split strategies and evaluate their consistency with realistic drug split scheme. 

\mysubsection{Drug Distribution Change as Surrogate}


In order to better capture distribution changes in realistic drug development process and conduct emerging DDI prediction evaluation,
we propose a distribution change simulation framework~(Figure~\ref{fig:framework_comp}(b)), where we use drug distribution changes as a surrogate to simulate the distribution changes in emerging DDI prediction problem.
It mainly contains steps as follows:

\begin{itemize}
\item \textbf{Drug distribution change modeling}: Introduce 
a measurement to model the distribution change between known drug set $D_k$ and new drug set $D_n$.
Use that distribution change as a surrogate of distribution change between training and test DDIs in emerging DDI prediction problem. 
\item \textbf{Drug data split}: Conduct a split of drugs into known and new drug sets, then obtain DDI dataset for emerging DDI prediction evaluation.
\end{itemize}

For common DDI data split~(as shown in Figure~\ref{fig:framework_comp}(a)) in most existing works
(e.g.~\citep{liu2022predict, yao2022effective, zitnik2018modeling, zhang2023emerging}), 
the phenomenon of distribution change is not considered, leading to unrealistic DDI evaluation results.
Through the distribution change simulation framework, we can better simulate the distribution changes and conduct emerging DDI prediction evaluation in a scene closer to real-world scenarios.

\mysubsection{Understanding}

\begin{figure}[t]
	\centering
	\includegraphics[width=0.48\textwidth]{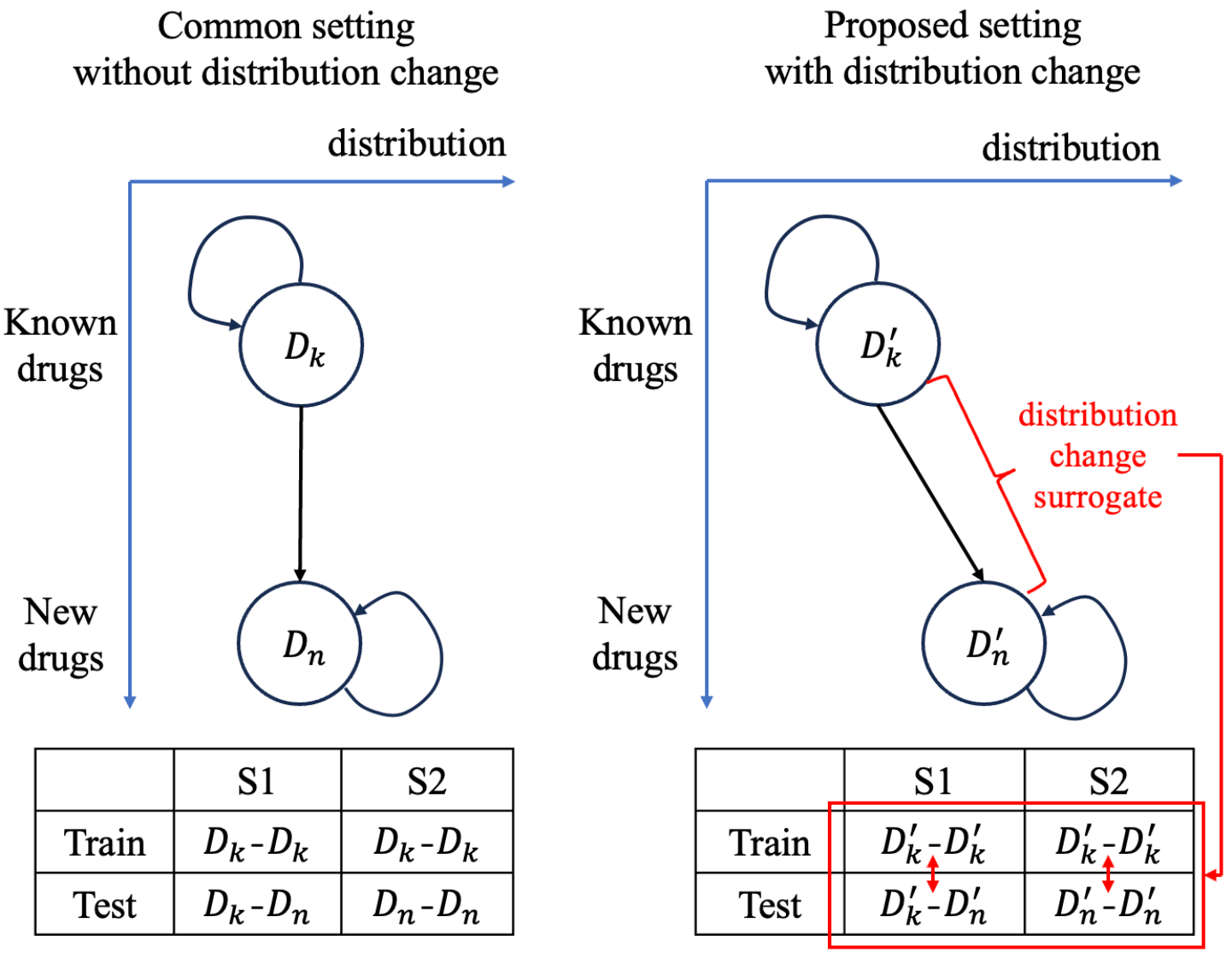}
	\vspace{-15px}
	\caption{Illustration of DDI data split.}
	\label{fig:DDIsp}
	   \vspace{-15px}
\end{figure}

The split of DDI data in different settings in S1 and S2 tasks is illustrated in Figure~\ref{fig:DDIsp}. 
In common setting, there is no distribution change 
between known drug set $D_k$ and new drug set $D_n$. 
The training and test DDIs for S1 and S2 tasks follow the same distribution, which does not conform with real-world drug development process.
Compared with common setting, 
the proposed setting introduces the distribution change between known drug set $D_k'$ and new drug set $D_n'$ as a surrogate of the distribution changes between training and test DDIs for emerging DDI prediction evaluation.
By incorporating these two different evaluation settings, 
we can more effectively benchmark the impact of distributional changes on DDI prediction methods under realistic scenarios.

\mysubsection{Customized Distribution Change Surrogate}
\label{sec:cluster}

\begin{figure}[t]
    \centering
    \includegraphics[width=0.45\textwidth]{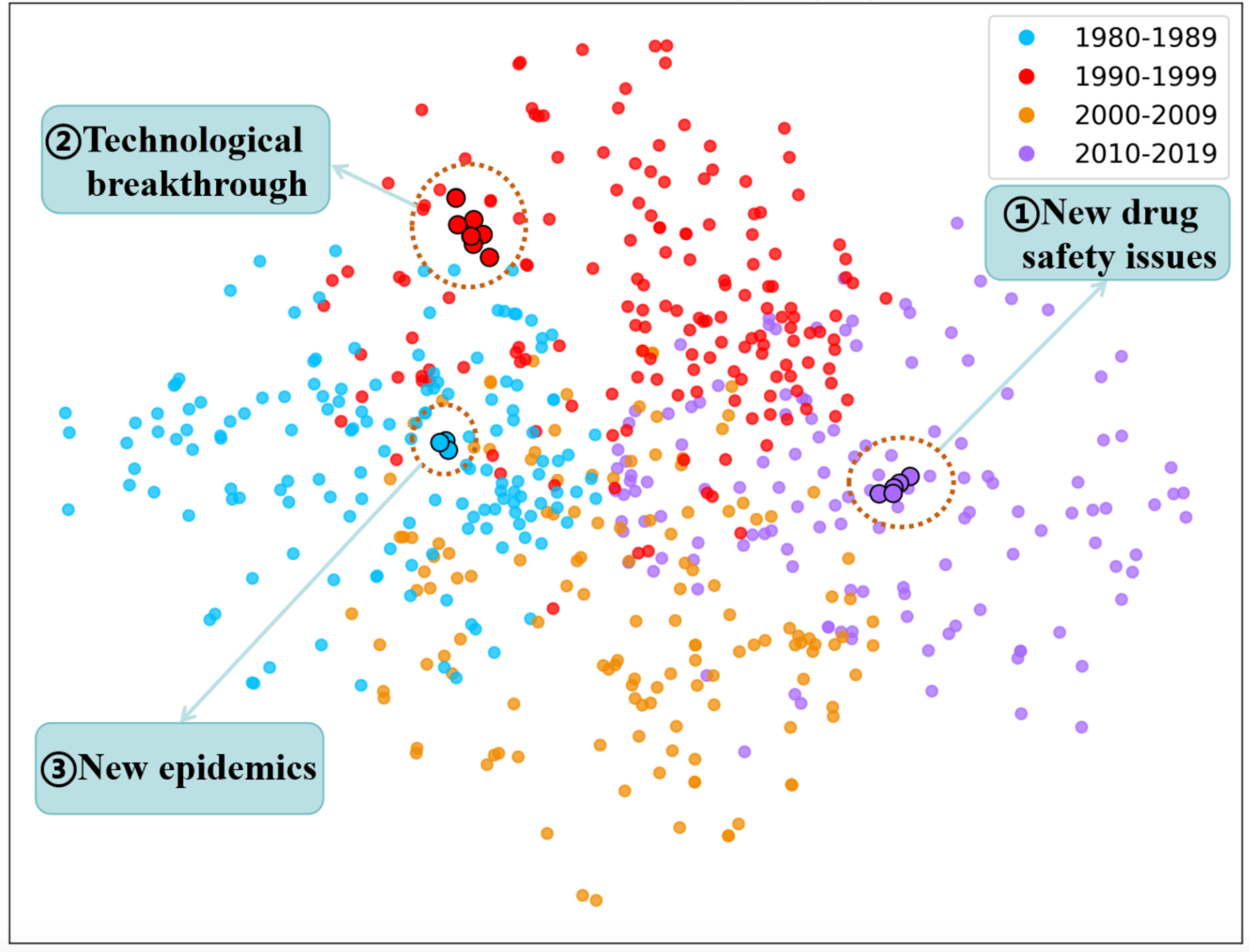}
    \caption{T-SNE illustration of drug distribution with their approval time in Drugbank dataset.}
    \label{fig:ill_time}
   \vspace{-15px}
\end{figure}

Following the above distribution change simulation framework, we first collect approximate approval times of a part of drugs in Drugbank dataset and visualize their distribution in the chemical space~(as shown in Figure~\ref{fig:ill_time}). 
We can see that drugs developed in specific time periods demonstrate a clustering effect in the chemical space owing to various factors, such as (1) new drug safety issues~\citep{rao2008evolution}, (2) technological breakthrough for drug development~\citep{cushman1991history}, and (3) new epidemics~\citep{pawlotsky2013ns5a}. 

Based on the above observation, 
we consider to design a customized cluster-based difference measurement to model the distribution changes between known and new drug set. 
Denote known and new drug set as $D_k$ and $D_n$.
We define difference between two drug sets as $\gamma(D_k, D_n) = max\{S(u, v),  \forall u \in D_k, v \in D_n\}$, where $S(\cdot, \cdot)$ is similarity measurement of two drugs.
Then we utilize parameter $\gamma$ as a surrogate to control distribution changes between training and test DDIs in emerging DDI prediction evaluation.
With the decrease of $\gamma$, the difference between known and new drug set will become more significant, leading to larger distribution changes between training and test DDIs.

\vspace{-10px}

\begin{remark} \it
The proposed cluster-based difference measurement is a reasonable surrogate under the proposed distribution change simulation framework. 
Additional factors that may lead to distributional changes between known and new drug sets could also be simulated and explored in future studies. 
\end{remark}

\vspace{-10px}

\begin{table*}[ht]
	\caption{Comparison among different data split strategies. }
	\small
	\vspace{-5px}
	\label{tab:split_comp}
    \setlength\tabcolsep{5pt}
	\begin{center}
		\begin{tabular}{ccccccccc}
			\hline
			Split strategy & Random & Drug frequence & Drug usage & Scaffold & LoHi & SPECTRA & DataSAIL & Cluster \\ \hline
            Preserve all data & \checkmark &\checkmark&\checkmark&\checkmark&\texttimes&\texttimes&\checkmark&\checkmark \\
            Controllable distribution change &\texttimes&\texttimes&\texttimes&\texttimes&\texttimes&\checkmark&\texttimes&\checkmark \\
            Consistency with approval time & Low & Low & Low & Low & Medium & Low & Medium & High \\
			\hline
		\end{tabular}
	\end{center}
	\vspace{-10px}
\end{table*}

\begin{figure}[t]
    \centering
    \includegraphics[width=0.45\textwidth]{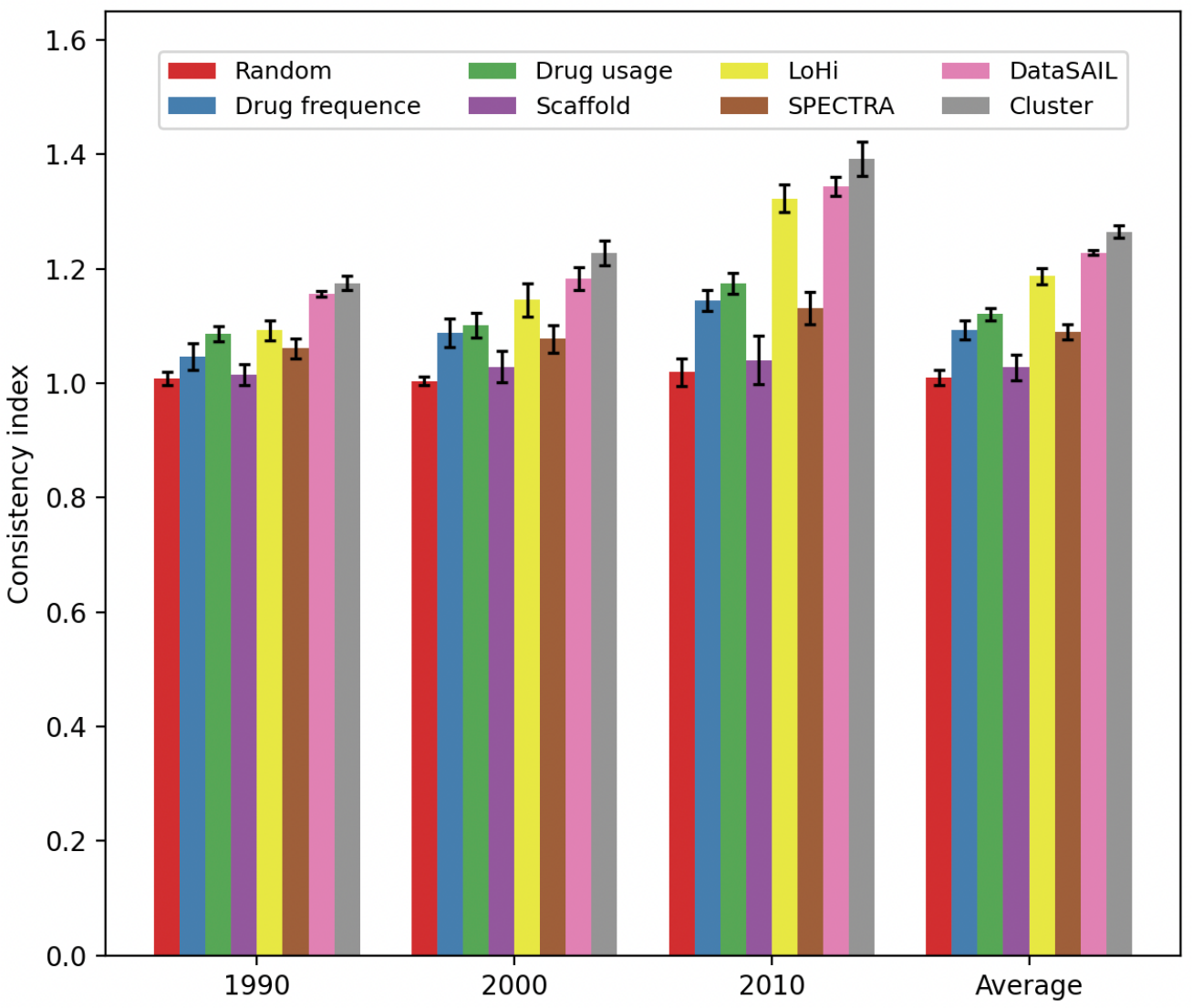}
	\vspace{-8px}
    \caption{Figure for time threshold of the realistic drug split scheme w.r.t the consistency index of each drug split scheme.}
    \label{fig:consistency}
    \vspace{-15px}
\end{figure}

{

\mysubsection{Comparison among Different Data Split Strategies}
\label{sec:consistency}

In this section, we provide comparison among different data split strategies as follows: (1) random split~\citep{zitnik2018modeling}, (2) drug frequency based split~\citep{jamal2020rethinking}, (3) drug usage based split~\citep{glorot2011domain}, (4) scaffold split~\citep{yang2019analyzing} (5) LoHi~\citep{steshin2023hi}, (6) SPECTRA~\citep{ektefaie2024evaluating}, (7) DataSAIL~\citep{joeres2025data} and (8) the proposed cluster based split~\citep{koh2021wilds}. 

Table~\ref{tab:split_comp} shows the overall comparison among the eight data split strategies. 
LoHi and SPECTRA need to discard a part of data points in the split process, which is generally viewed as undesirable given the high cost of drug data collection.
Only SPECTRA and the cluster-based split generate splits that involve controllable distributional changes, enabling the examination of how different DDI prediction methods perform under increasingly pronounced distribution changes.


To further assess the consistency between different split strategies and realistic data, we conducted verification experiments.
Using the recorded approval times of a subset of drugs from the DrugBank dataset, we designated specific time points as thresholds to separate drugs into known and new categories, thereby constructing a realistic drug split scheme.
For each threshold, we computed a consistency index between every drug split scheme and the realistic split, quantifying the degree of alignment for drug split strategies with real-world scenarios.
The results, presented in Figure~\ref{fig:consistency}, demonstrate that the proposed cluster-based split consistently achieves the highest consistency index, indicating closer conformity to the realistic split. 
Compared with other split strategies, LoHi and DataSAIL are also relatively more consistent with realistic drug split scheme.
Based on these results, we choose cluster-based split to simulate realistic distribution changes in benchmarking evaluation of emerging DDI prediction. 

}

\vspace{-15px}

\section{Instantiation of the Proposed Framework}

In this section, we first instantiate the proposed distribution change simulation framework by drug split 
based on cluster-based difference measurement for emerging DDI prediction evaluation. 
The evaluated methods and evaluation metrics in benchmarking experiments are subsequently introduced. 

\mysubsection{Dataset and Split}
\label{sec:dataset}
In DDI-Ben, we conduct experiments on two widely used public DDI datasets: 
(1) Drugbank~\citep{wishart2018drugbank}, a multiclass DDI prediction dataset where 
each drug pair in this dataset is associated with one of 86 possible interaction types.
(2) TWOSIDES~\citep{tatonetti2012data}, a multilabel DDI prediction dataset that records side effects between drugs. 
We keep 209 DDI types~(occurrence frequency from 3000 to 6000) to ensure each DDI type corresponds to enough drug pairs for learning, 
and each drug pair may have multiple interactions among 209 DDI types.

The drug split based on cluster-based difference measurement is shown in Algorithm~\ref{alg:split}.
Through the algorithm, the difference measurement between known and new drug set satisfies $\gamma(D_k, D_n) \leq \gamma_0$. 
Hence, we can control the distribution changes between known and new drug sets through adjustment of the parameter $\gamma_0$, where smaller $\gamma_0$ indicates more significant distribution changes. 
For common DDI evaluation setting, we randomly split drugs into known drug set $D_k$ and new drug set $D_n$.
With known and new drug sets, we split DDI data for emerging DDI prediction evaluation.
In both S1 and S2 tasks, training set includes all DDI triplets $(u, r, v)$ where $u$ and $v$ are both in known drug set $D_k$. 
The test set in S1 task comprises DDI triplets $(u, r, v)$ where one drug $u$ is in known drug set  $D_k$ and the other drug belongs to new drug set $D_n$.
The test set in S2 task contains DDI triplets $(u, r, v)$ where both drugs $u$ and $v$ are in new drug set $D_n$.

\begin{algorithm}[t]
	\caption{Drug split based on cluster based difference measurement.}
	\small
	\begin{algorithmic}[1]
		\Require The drug set $D$, the similarity measurement $S(\cdot, \cdot)$.
		\State Calculate the similarity between drug pairs in the dataset $D$ based on measurement $S$. 
		\State Determine the distribution change parameter $\gamma_0$. Regard different drugs as nodes in a graph. For each drug pair in the dataset, if their similarity is larger than $\gamma_0$, we build connection between these two drugs. 
		\State Based on connections of drugs in the graph, we regard each connected component in the graph as a drug cluster. The drug graph is divided into $m$ clusters $D_1, D_2, ..., D_m$.
		\State Randomly divide drug clusters into known drug set $D_k$ and new drug set $D_n$.
		\Return Known drug set $D_k$ and new drug set $D_n$.
	\end{algorithmic}
	\label{alg:split}
\end{algorithm}

\mysubsection{Methods to be Compared}

{

We choose ten representative methods to be compared in DDI-Ben. 
Their category and used side information are summarized in Table~\ref{tab:prior}.
MLP~\citep{rogers2010extended} is the most classical feature based method that designs a multi-layer perceptron to predict the DDI based on input drug fingerprints.
MSTE~\citep{yao2022effective} specially designs a knowledge graph embedding scoring function that can perform well for DDI prediction problem.
Decagon~\citep{zitnik2018modeling} is representative GNN based methods that consider to incorporate biomedical network into DDI prediction. 
SSI-DDI~\citep{nyamabo2021ssi} designs a GNN to model drug molecular graphs and predict DDI based on interaction between substructure of query drug pairs. 
MRCGNN~\citep{xiong2023multi} presents a multi-relation graph contrastive learning strategy to better characteristics of rare DDI types based on drug molecular structures. 
EmerGNN~\citep{zhang2023emerging} is a GNN based method that specially designs a flow based graph neural network to predict emerging DDIs.
SAGAN~\citep{zhang2025domain} utilizes a transfer learning strategy to enhance the cross-domain generalization ability of GNNs on DDI prediction tasks. 
TIGER~\citep{su2024dual} is a graph-transformer based method that designs a dual chanel graph transformer to both capture the structural information of drug molecular graphs and the information from biomedical network.
TextDDI~\citep{zhu2023learning} is a LLM based method that first try to use large language model to predict DDI based on textual information of drugs and drug interactions.
DDI-GPT~\citep{xu2024ddi} captures relevant information of query drugs from biomedical networks and uses biomedical large language model to enhance DDI prediction performance. 
}

\begin{table*}[ht]
	\caption{Main characteristics about DDI prediction methods we compared. }
	\small
	\vspace{-8px}
	\label{tab:prior}
    \setlength\tabcolsep{7pt}
	\begin{center}
		\begin{tabular}{ccc}
			\toprule
			Method&Categorization&Side information used\\
			\midrule
			MLP~\citep{rogers2010extended}&Feature based&Drug fingerprints\\
			MSTE~\citep{yao2022effective}&Embedding based&-\\
			Decagon~\citep{zitnik2018modeling}&GNN based&Drug fingerprints, biomedical networks\\
            SSI-DDI~\citep{nyamabo2021ssi}&GNN based& Drug structural information\\
            MRCGNN~\citep{xiong2023multi}&GNN based& Drug structural information\\
			EmerGNN~\citep{zhang2023emerging}&GNN based&Drug fingerprints, biomedical networks\\
            SAGAN~\citep{zhang2025domain}&GNN based& Drug structural information\\
			TIGER~\citep{su2024dual}&Graph-transformer based&Drug structural information, biomedical networks\\
			TextDDI~\citep{zhu2023learning}&LLM based&Textual data of drugs and drug interactions\\
            DDI-GPT~\citep{xu2024ddi}&LLM based& Textual data of drugs and drug interactions, biomedical networks\\
      \midrule
	\end{tabular}
	\end{center}
	\vspace{-20px}
\end{table*}

\mysubsection{Evaluation Metric}
\label{sec:metric}

Following the common practices~\citep{yu2021sumgnn,zhang2023emerging}, we use F1~(primary), accuracy, Cohen's Kappa~\citep{cohen1960coefficient} as evaluation metrics in multiclass DDI prediction for Drugbank dataset.
Following the evaluation of~\citep{tatonetti2012data, zitnik2018modeling} in multilabel DDI prediction for TWOSIDES dataset, 
we report the average of ROC-AUC~(primary), PR-AUC, and accuracy in each DDI type. 
The experimental results for evaluation metric accuracy, Cohen's Kappa on Drugbank and PR-AUC, accuracy on TWOSIDES are shown in supplementary material.

\section{Empirical Results}
\label{sec:exp}

We conduct extensive experiments to evaluate emerging DDI prediction in the proposed DDI-Ben benchmark in this section. 
Through experiments in settings with and without distribution changes, it is observed that most of existing DDI prediction methods have significant performance drop with distribution changes introduced, and LLM based methods are the most robust type of method against this negative impact. 
We further analyze performance on different DDI types and control distribution changes through parameter $\gamma$, providing in-depth insights into the influence of distribution changes.

\begin{figure*}[t]
	\centering
	\includegraphics[width=0.6\textwidth]{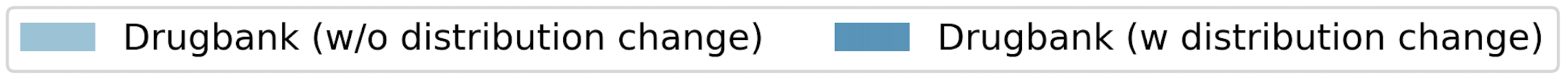}
	\vspace{-5px}
	
	\subfigure[\normalsize S1 task on Drugbank. ]
	{\includegraphics[width=0.48\textwidth]{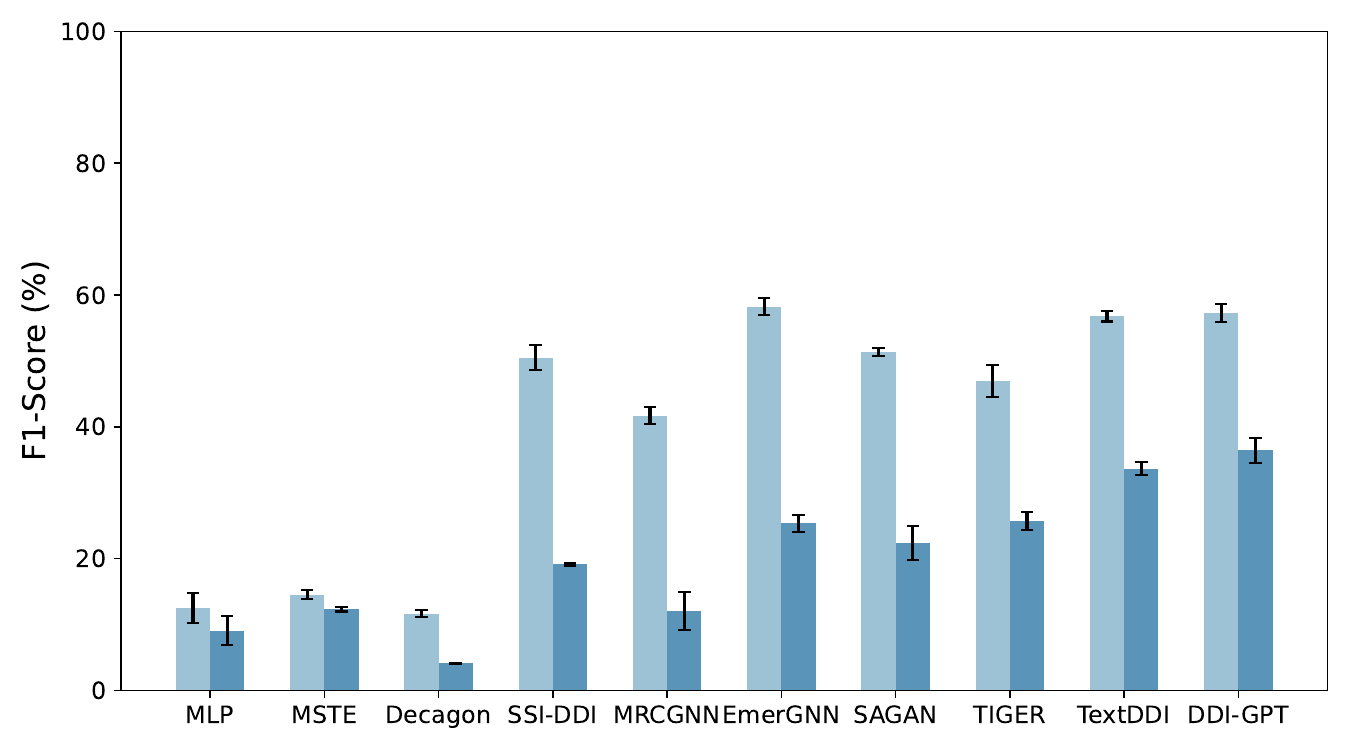}}
    \qquad
	\subfigure[\normalsize S2 task on Drugbank. ]
	{\includegraphics[width=0.48\textwidth]{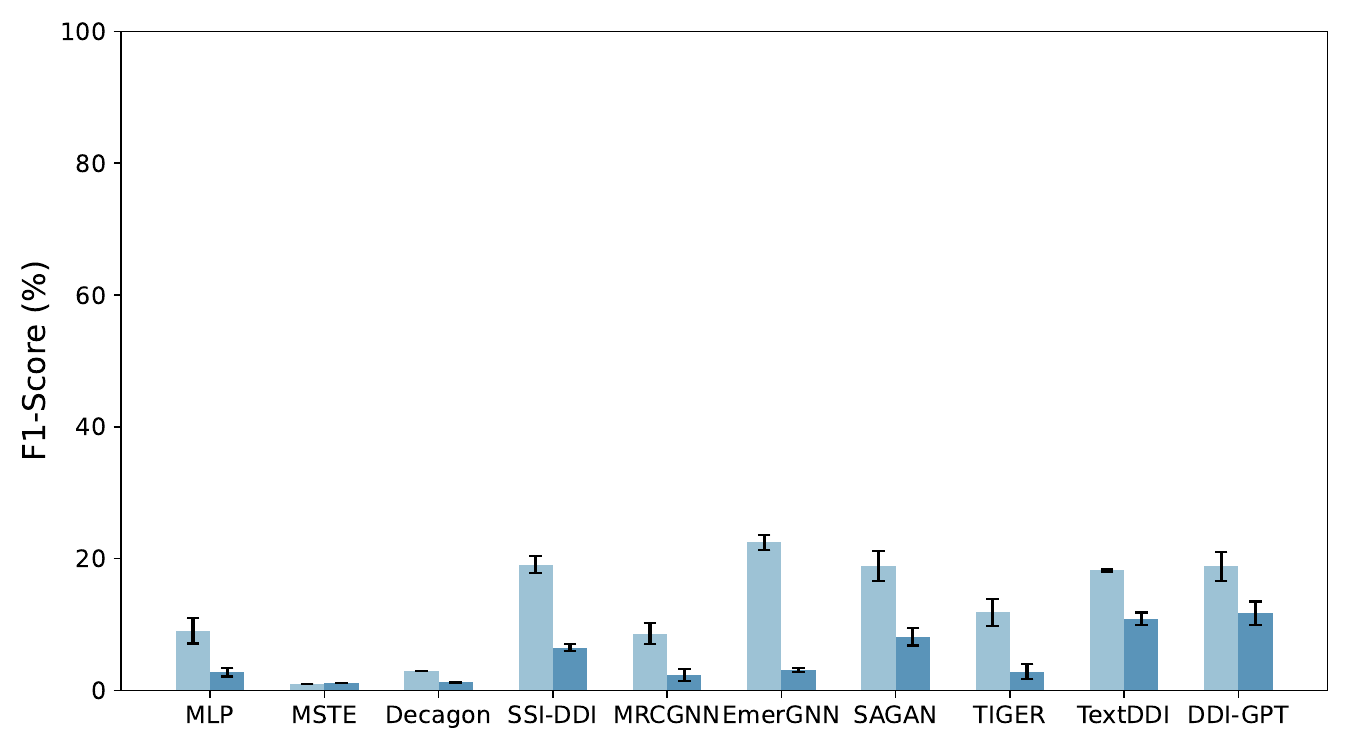}}
	
	\includegraphics[width=0.6\textwidth]{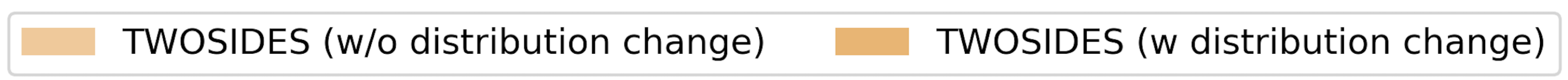}
	\vspace{-5px}
	
	\subfigure[\normalsize S1 task on TWOSIDES. ]
	{\includegraphics[width=0.48\textwidth]{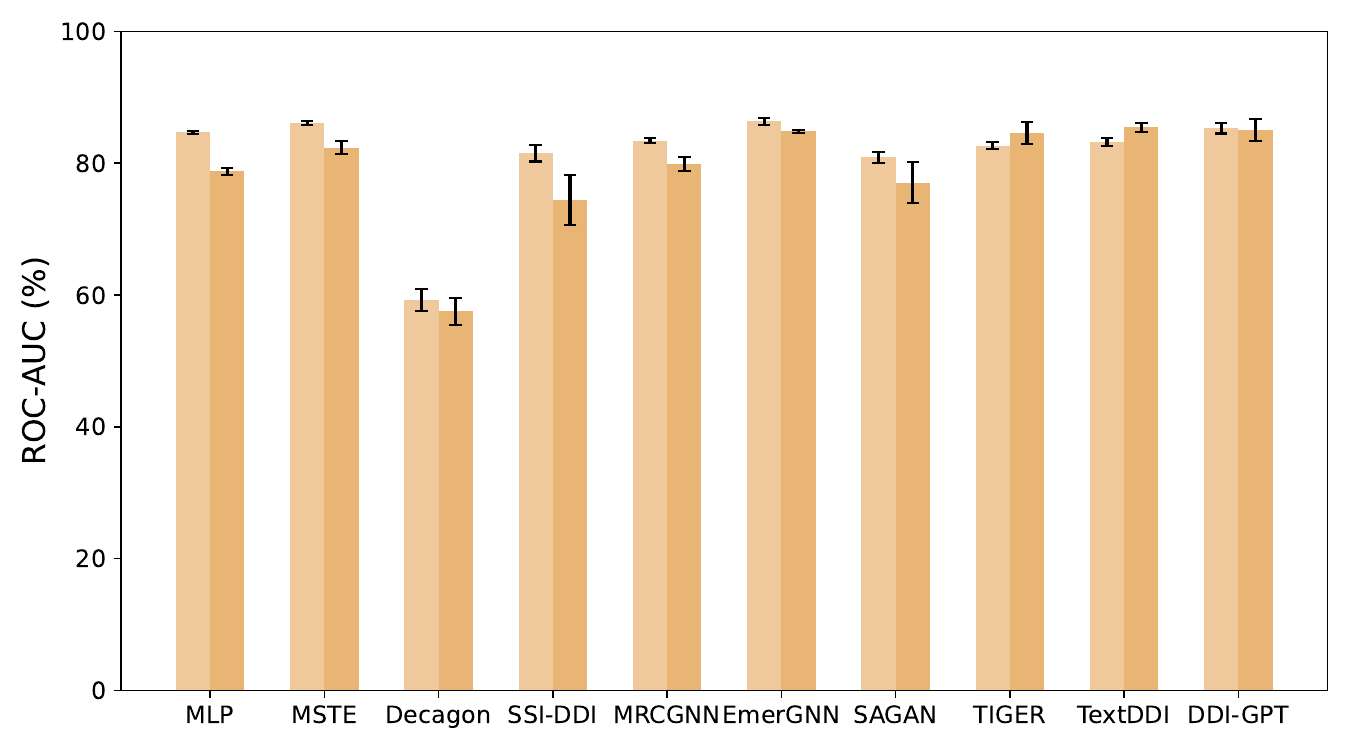}}
    \qquad
	\subfigure[\normalsize S2 task on TWOSIDES. ]
	{\includegraphics[width=0.48\textwidth]{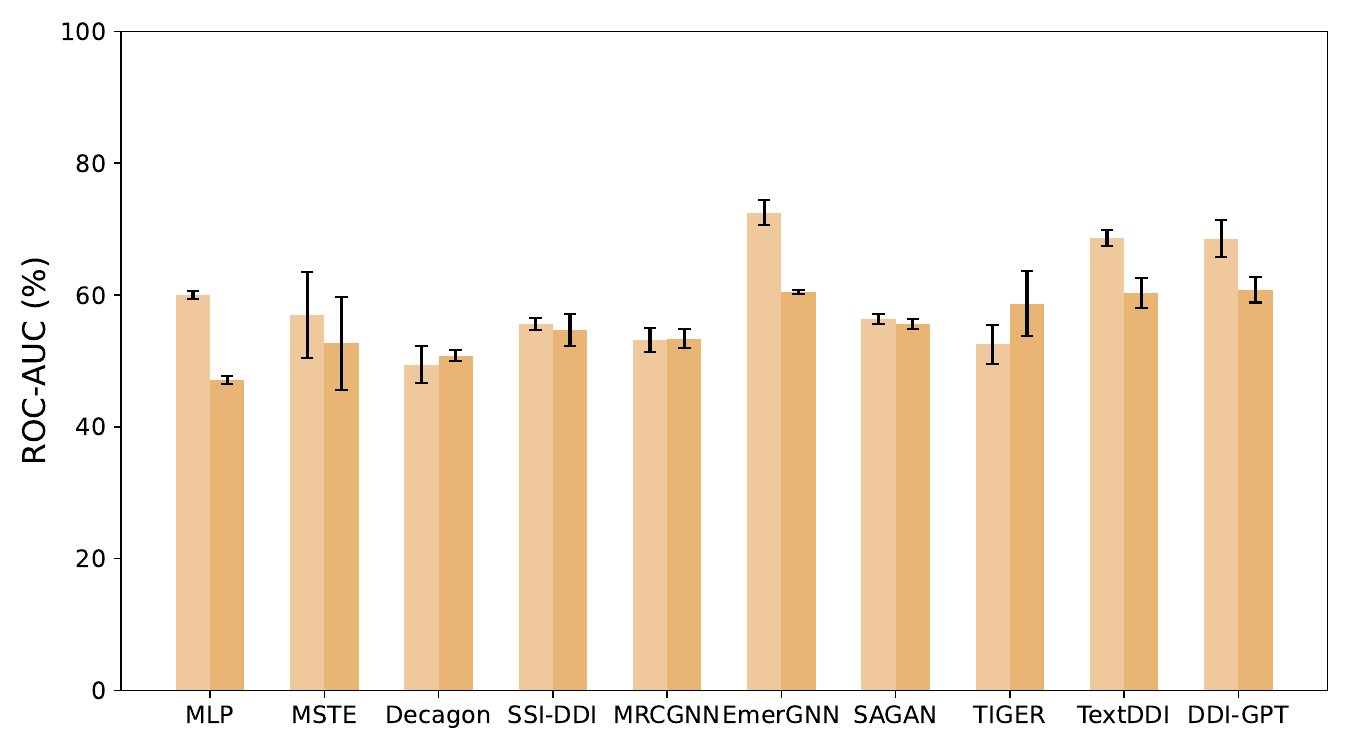}}
	\caption{{Performance comparison for different types of DDI methods in the settings with and without distribution change in S1-S2 tasks. 
		Here we utilize primary evaluation metric for each dataset~
		(F1 for Drugbank and ROC-AUC for TWOSIDES).}}
	\label{fig:overview_f}
	\vspace{-5px}
\end{figure*}

\mysubsection{Performance Analysis among Representative Methods}
\label{sec:exp_overall}

We first conduct experiments on the two datasets with and without distribution changes introduced, and compare the performance of different DDI methods in two emerging DDI prediction tasks~(S1-S2).
The results are shown in Figure~\ref{fig:overview_f}.

For general comparison, we can see that injecting prior knowledge into method design may improve the performance. 
GNN-based and graph transformer based methods outperform feature based and embedding based approaches, as they better leverage structural information from DDI graphs and side information from biomedical networks through neighborhood propagation.
When comparing the two datasets, existing methods experience a more substantial performance decline on Drugbank when distribution changes are introduced, compared to their decline on TWOSIDES.
This discrepancy can be attributed to the higher difficulty of multiclass prediction in the Drugbank dataset, where the larger number of classes increases the likelihood of misclassification under distribution changes.

\begin{table*}[t]
	\caption{A case study on textual side information used and text process outcomes of TextDDI. 
		The descriptions in ``Raw textual data'' part with italic script contain pharmacological knowledge. The words marked in orange  are important information relevant to DDI prediction of the two query drugs. }
	\scriptsize
    \vspace{-10px}
	\label{tab:textddi}
	\begin{center}
		\begin{tabular}{m{1.5cm}|m{15.7cm}}
			\toprule
			Raw textual data  & \textbf{Aclidinium}: Aclidinium does not prolong the QTc interval or have significant effects on cardiac rhythm. Aclidinium bromide inhalation powder is \emph{indicated for the long-term, maintenance treatment of bronchospasm associated with chronic obstructive pulmonary disease~(COPD)}, including chronic bronchitis and emphysema. It \emph{has a much higher propensity to bind to muscarinic receptors than nicotinic receptors}. FDA approved on July 24, 2012. Prevention of \textcolor{orange}{acetylcholine-induced} bronchoconstriction effects was dose-dependent and lasted longer than 24 hours. \textbf{Butylscopolamine}: Used to treat abdominal cramping and pain. Scopolamine butylbromide \emph{binds to \textcolor{orange}{muscarinic M3 receptors}} in the gastrointestinal tract. The inhibition of contraction reduces spasms and their related pain during abdominal cramping. 
			\textbf{Prediction}: We can predict that the drug-drug interaction between Aclidinium and Butylscopolamine is that: Aclidinium may \textbf{\textcolor{red}{decrease the bronchodilatory activities}} of Butylscopolamine.\\
			\midrule
			Prompt processed by TextDDI&
			\textbf{Aclidinium}: Prevention of \textcolor{orange}{acetylcholine-induced} bronchoconstriction effects was dose-dependent and lasted longer than 24 hours. Aclidinium is a long-acting, competitive, and reversible \textcolor{orange}{anticholinergic} drug that is specific for the \textcolor{orange}{acetylcholine muscarinic} receptors. It binds to all 5 muscarinic receptor subtypes to a similar affinity. Aclidinium's effects on the airways are mediated through the M3 receptor at the smooth muscle to cause bronchodilation. \textbf{Butylscopolamine}: This prevents \textcolor{orange}{acetylcholine} from binding to and activating the receptors which would result in contraction of the smooth muscle. Scopolamine butylbromide binds to muscarinic M3 receptors in the gastrointestinal tract. The inhibition of contraction reduces spasms and their related pain during abdominal cramping. \textbf{Prediction}: We can predict that the drug-drug interaction between Aclidinium and Butylscopolamine is that: Aclidinium may \textbf{\textcolor{darkgreen}{increase the anticholinergic activities}} of Butylscopolamine.\\
			\midrule
		\end{tabular}
	\end{center}
	\vspace{-20px}
\end{table*}

\begin{table*}[t]
	\caption{DDI prediction performance for different DDI types on Drugbank~(S1 task). Here ``Major'', ``Medium'', ``Long-tail'' denote DDI types with high, medium, low occurrence frequency, respectively.
    ``w/o'' and ``w'' denote the setting without and with distribution change introduced.
    For each DDI type, the best results for ``w/o'' and ``w'' setting are marked by \underline{underline} and \textbf{bold}, respectively.}
	\footnotesize
	\vspace{-8px}
	\label{tab:type}
    \setlength\tabcolsep{8pt}
	\begin{center}
		\begin{tabular}{cc|ccc|ccc|ccc}
			\hline
           \multirow{2}{*}{Method} & Distribution & \multicolumn{3}{c|}{Major} & \multicolumn{3}{c|}{Medium} & \multicolumn{3}{c}{Long-tail}\\
            & change & \#48 & \#46 & \#72 & \#29 & \#71 & \#57 & \#24 & \#1 & \#18 \\ \hline
            \multirow{2}{*}{MLP} &w/o & \underline{84.9} & 56.9 & \underline{64.2} & 25.6 & 64.1 & 32.6 & 27.1 & 0.0 & 0.0 \\
             &w & 73.9 & 52.3 & \textbf{44.5} & 0 & 17.5 & 13.2 & 15.1 & 0.0 & 0.0 \\ \hline
            \multirow{2}{*}{MSTE} &w/o & 80.0 & 65.5 & 52.3 & 16.4 & 24.0 & 30.4 & 0.0 & 22.2 & 0.0 \\ 
             &w & 74.9 & \textbf{63.2} & 37.9 & 2.9 & 10.9 & 0.3 & 9.9 & 17.1 & 0.0 \\ \hline
            \multirow{2}{*}{EmerGNN} &w/o & 82.4 & \underline{73.0} & 59.3 & 77.2 & \underline{96.7} & 86.0 & \underline{59.1} & \underline{83.3} & \underline{75.0} \\
             &w & 65.3 & 59.1 & 40.8 & 61.8 & 24.1 & 56.7 & 35.1 & 55.7 & 35.0 \\ \hline
            \multirow{2}{*}{TIGER} &w/o & 75.8 & 60.4 & 55.4 & 46.7 & 93.8 & \underline{95.9} & 27.6 & 53.7 & 62.5 \\
             &w & 73.5 & 42.5 & 37.9 & 32.3 & 56.1 & \textbf{87.4} & 5.6 & 48.5 & 13.3 \\ \hline
            \multirow{2}{*}{DDI-GPT} &w/o & 82.5 & 69.5 & 50.7 & \underline{80.7} & 92.4 & 89.2 & 54.4 & 66.7 & 65.0 \\
             &w & \textbf{76.1} & 62.3 & 42.3 & \textbf{67.7} & \textbf{72.4} & 84.3 & \textbf{58.3} & \textbf{61.4} & \textbf{53.3} \\ \hline
		\end{tabular}
	\end{center}
	\vspace{-10px}
\end{table*}

\mysubsection{Revealing Factors that can Alleviate the Negative Impact of Distribution Change}

From the experimental results in Figure~\ref{fig:overview_f}, 
it can be seen that almost all of the evaluated methods exhibit performance degradation under distribution changes.
However, the extent of performance degradation varies. 
Specifically, EmerGNN achieves the highest performance in the existing setting without distribution changes, whereas TextDDI and DDI-GPT outperform other methods in scenarios involving distribution changes. 
To further analyze the reason for that result, we conduct a case study for the textual side information used and text processing outcomes of TextDDI~\citep{zhu2023learning} in Table~\ref{tab:textddi}. 
TextDDI's raw data include crucial drug information, such as names and interactions, adding realistic semantics to DDI prediction. 
The textual descriptions of two query drugs contain transferable pharmacological knowledge, 
like Aclidinium's bronchodilatory activities and Butylscopolamine's treatment and related receptors. 
This knowledge is not provided in other side information types, like drug fingerprints and biomedical networks. 
However, the DDI classifier struggles with irrelevant and distracting information in the raw data. 
The LLM that TextDDI uses effectively addresses this issue with its strong text comprehension and processing abilities.
The experimental results show that LLM model in TextDDI obtains key pharmacological knowledge through textual side information, filter out irrelevant information, and provide useful data for DDI prediction. With TextDDI's processed prompt, the DDI classifier can make accurate predictions.

\mysubsection{Performance Analysis for Different DDI Types}
\label{sec:relation}

Different DDI types can vary widely in their mechanism and they may be affected differently by realistic distribution changes in DDI prediction. 
Here we conduct experiments to analyze 
the performance of DDI prediction methods for different DDI types.
We select the best-performed methods from five DDI prediction categories as representatives, including MLP, MSTE, EmerGNN, TIGER, and DDI-GPT. 
Part of results on S1 task on Drugbank dataset are shown in Table~\ref{tab:type}, where we sample 3 major DDI types, 3 DDI types with medium frequency, and 3 long-tail DDI types.

Generally, MLP and MSTE can achieve good performance on major DDI types, but they perform poorly on medium and long-tail DDI types.
This suggests that these methods largely rely on sufficient labelled training data for prediction.
Compared with two methods above, EmerGNN, TIGER and DDI-GPT can achieve better performance on medium and long-tail DDI types.
When distribution changes are introduced, EmerGNN and TIGER have a significant performance drop on all of DDI types. 
DDI-GPT significantly outperforms other methods on medium and long-tail DDI types with distribution changes introduced, indicating that LLMs can effectively handle the problem of distribution changes and lack of labelled training data for emerging DDI prediction.

\mysubsection{Controlling Distribution Change}
\label{sec:gamma}

Here we tune the parameter $\gamma$~(introduced in Section~\ref{sec:cluster}) to control distribution changes between drug sets and analyze the performance of existing DDI methods with different normalized $\gamma$ values.
Similar as Section~\ref{sec:relation}, we evaluate representative methods from each of the five categories. 

\begin{figure*}[t]
	\centering
	\includegraphics[width=0.5\textwidth]{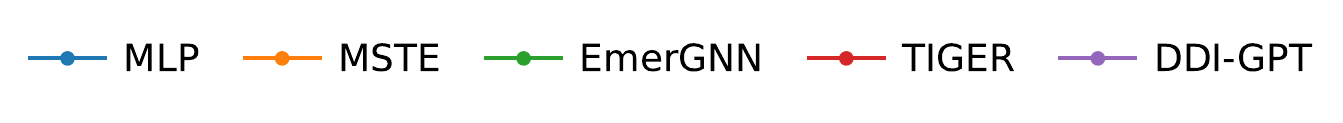}
	
	\vspace{-15px}
	
	\subfigure[\normalsize S1 task on Drugbank. ]
	{\includegraphics[width=0.4\textwidth]{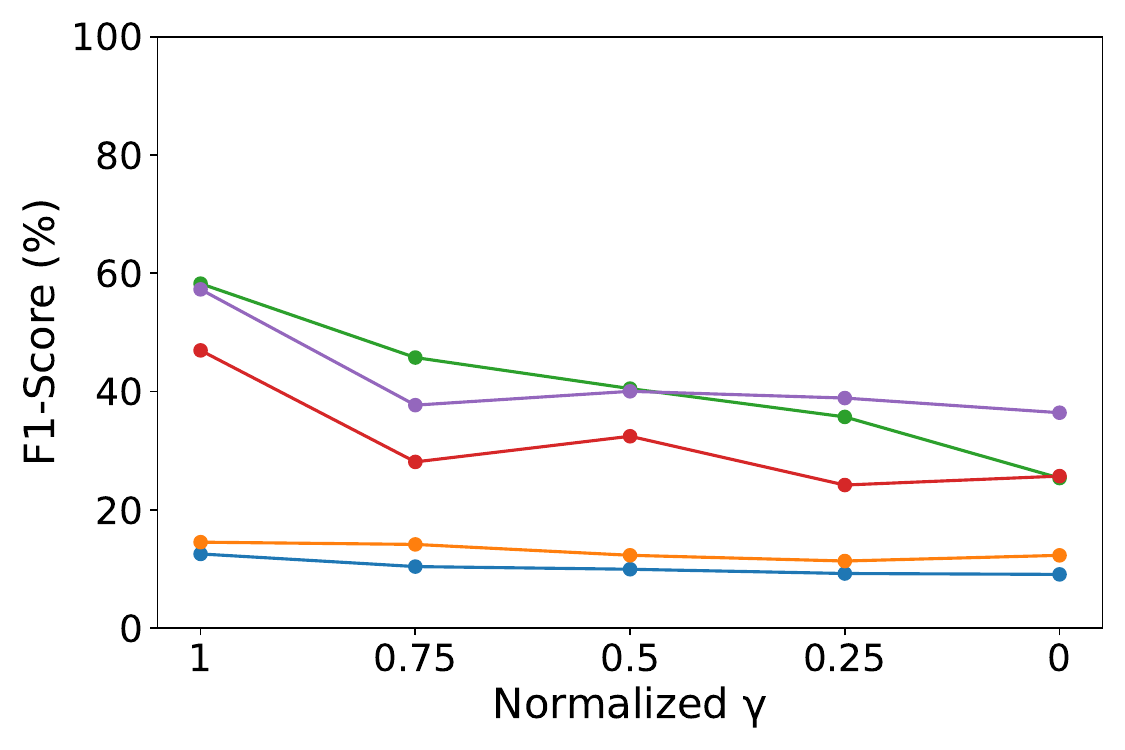}}
    \qquad
	\subfigure[\normalsize S2 task on Drugbank. ]
	{\includegraphics[width=0.4\textwidth]{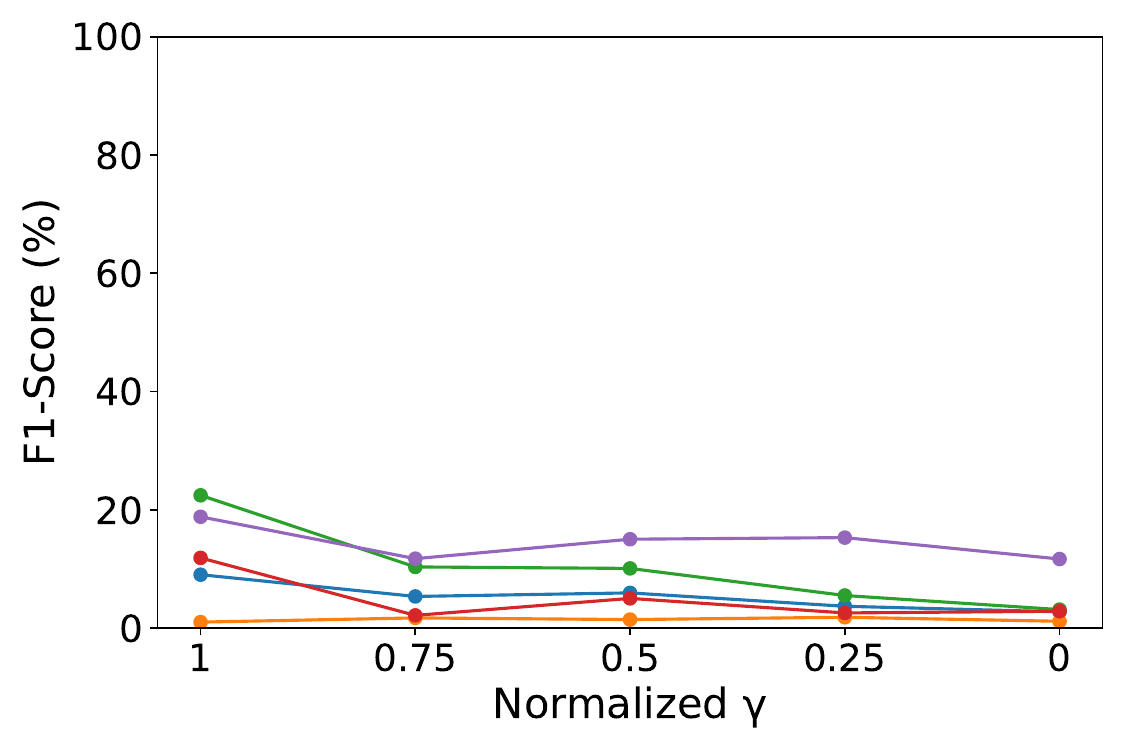}}

	\vspace{-5px}
	\caption{{Tuning $\gamma$ in the setting with distribution change for DDI prediction on Drugbank dataset. }}
	\vspace{-15px}
	\label{fig:gamma_f1}
\end{figure*}

The experiment results on Drugbank dataset are demonstrated in 
Figure~\ref{fig:gamma_f1}.
With the decrease of the parameter $\gamma$, the distribution changes between training and test DDIs also become more significant.
The performances of MLP and MSTE are unsatisfactory compared with other methods.
For the other three methods, 
their performance drops significantly with distribution change becoming more significant, suggesting their sensitivity to distribution changes.
Also, among the existing DDI methods, DDI-GPT performs relatively better than other methods when distribution changes are more significant.
This suggests that the prior knowledge in LLMs and the textual side information used in DDI-GPT can help alleviate the negative impact of distribution changes on DDI prediction performance.

\section{Related Works}
\label{sec:related}

\noindent
\textbf{Drug-Drug Interaction Prediction.}
The field of drug-drug interaction (DDI) prediction has attracted increasing attention in recent years and different types of techniques have been proposed.
Feature based methods~\citep{rogers2010extended, ryu2018deep, liu2022predict} mainly rely on drug features and design a classifer to map the features of drug pairs to DDI types. 
Since data of drug interactions can be naturally represented as a graph, graph learning methods have been widely used to predict DDIs. 
Embedding based methods~\citep{karim2019drug, yao2022effective} follow the idea of knowledge graph embedding learning, which design a model to learn embeddings of drugs and interactions and uses the embeddings to measure the interactions. 
GNN based methods~\citep{zitnik2018modeling, lin2021kgnn, yu2021sumgnn,  zhang2023emerging, wang2024accurate} utilize graph neural networks to encode structure information from DDI or molecular graphs for DDI prediction.
Graph transformer based methods~\citep{su2024dual, chen2024drugdagt} introduce graph transformer to better capture global structural information from drug-relatd structural knowledge.
Large language model~(LLM) based methods~\citep{zhu2023learning, xu2024ddi} utilize a large language model to predict DDI based on descriptions of drugs and relations.

\noindent
\textbf{Simulation of Distribution Changes in Machine Learning.}
Distribution changes commonly exist when deep learning methods are used in real-world scenarios, where the training and test data follow different distributions. 
They typically stem from factors such as covariate shift in input features, sampling bias in data collection, or concept drift in dynamic systems. 
To systematically study this phenomenon, existing works often simulate distribution changes. 
For example, in computer vision, it is common to use images captured under varied conditions as source and target datasets, thereby simulating distribution shifts encountered in real-world scenarios~\citep{glorot2011domain}.
There are also works that sample long-tail data, so as to simulate the distribution changes~\citep{jamal2020rethinking}.
Recently, strategies that simulate distribution changes have been introduced into bioinformatics to better capture realistic scenarios.
Lo-Hi~\citep{steshin2023hi} proposes drug split strategies to simulate lead optimization and hit identification tasks in real drug discovery process.
SPECTRA~\citep{ektefaie2024evaluating} evaluates model generalizability by plotting performance against decreasing cross-split overlap and reporting the area under this curve.
DataSAIL~\citep{joeres2025data} models data split as a combinatorial optimization problem and find leakage-reduced data splits based on optimization strategies.

\section{Conclusion and Future Works}

In this work, we propose DDI-Ben, a benchmark for DDI prediction in a perspective from drug distribution changes. 
We propose a distribution change simulation framework that is compatible with different drug split strategies to reflect distribution changes in real-world emerging DDI prediction problems. 
Through extensive benchmarking evaluation, we find that the distribution changes can cause significant performance drops for existing DDI methods and LLM based methods exhibiting superior robustness. 
To facilitate future research, we release benchmark datasets that include simulated distribution changes.
For future work, designing LLM based methods can potentially alleviate the negative impact of distribution changes in emerging DDI prediction. 
Domain adaptation methods can also be a potential research direction for practical solutions to deal with distribution changes in real-world emerging DDI prediction.

\section*{Acknowledgments}
We extend our sincere gratitude to Yang Cao for his assistance in manuscript preparation. We also thank the anonymous reviewers for their insightful comments and constructive suggestions.

\section*{Funding}
Q. Yao is supported by National Natural Science Foundation of China (under Grant No.92270106) and Beijing Natural Science Foundation (under Grant No.4242039). 

\noindent Y. Zhang is supported by Guangdong Basic and Applied Basic Research Foundation (Under Grant 2025A1515010304), and Guangzhou Science and Technology Planning Project 2025A03J4491.

\bibliographystyle{apalike}
\bibliography{reference}

\clearpage

\onecolumn

\begin{appendices}

\section{Supplementary Information of DDI-Ben}

\subsection{Detailed Information of Drug Clusters in Real-world Data}
\label{app:add_drug_cluster}

In Table~\ref{tab:detail_drug_cluster}, we provide the detailed information of drug clusters in Figure~\ref{fig:ill_time}, including the drug names and explanations for the clusters.

\begin{table*}[h]
    \small
    \caption{Detailed information of drug clusters in Figure~\ref{fig:ill_time}.}
    \center
    \label{tab:detail_drug_cluster}
    \begin{tabular}{m{1.8cm}|m{4.7cm}|m{5cm}|m{4.5cm}}
        \toprule
        Time period & 1980-1989 & 1990-1999 & 2010-2019 \\
        \midrule
        Clustering reasons & new drug safety issues & technological breakthrough & new epidemics \\\midrule
        Drugs in the cluster & Ibuprofen, Flurbiprofen, Fenoprofen & Ramipril, Benazepril, Moexipril, Fosinopril, Captopril, Perindopril, Quinapril, Trandolapril & Daclatasvir, Ledipasvir, Elbasvir, Pibrentasvir, Velpatasvir \\\midrule
        Explanation & Severe gastrointestinal side effects of early NSAIDs (e.g., aspirin-induced ulcers) prompted the 1980s development of safer alternatives like Ibuprofen, Fenoprofen and Flurbiprofen, which inhibit COX enzymes to alleviate inflammation with reduced toxicity. & Breakthroughs in RAAS pathophysiology led to 1990s ACE inhibitors (Ramipril, Benazepril, Moexipril, Fosinopril, Captopril, Perindopril, Quinapril and Trandolapril) that suppress angiotensin II, clinically proven to protect hearts and kidneys through targeted 
        vascular regulation. & Hepatitis C virus (HCV)-induced liver cirrhosis and cancer drove the development of NS5A inhibitors, including Daclatasvir, Pibrentasvir, Ledipasvir, Velpatasvir, and Elbasvir to directly block viral replication. \\
        \midrule
    \end{tabular}
\end{table*}

\subsection{Similarity Measurement for Cluster-based Drug Split} 

In this work, we utilize Tanimoto Coefficient between the fingerprints of two drugs as the similarity measurement $S(\cdot, \cdot)$ of drug pairs:
\begin{equation}
    S(u,v) = \frac{f(u)^\mathsf{T}f(v)}{||f(u)||^2 + ||f(v)||^2 - f(u)^\mathsf{T}f(v)}
\end{equation}
where $u, v$ are drugs and $f(\cdot)$ is the fingerprint of a drug. 
Actually, the Tanimoto Coefficient is a widely used similarity measurement between drugs in pharmacy. 


\subsection{Calculation of Consistency Index in Section~\ref{sec:consistency}}
\label{app:consistency}

Here we present the calculation strategy of consistency index mentioned in Section~\ref{sec:consistency} in Algorithm~\ref{alg:consistency}.
Note that we denote the approval time of a certain drug $u$ as $y_u$. 
We use a penalty value $P_i$ to measure each error in drug split, and we 
assume that the larger the distance, the more severer the split error is.

\begin{algorithm}
    \caption{Calculation of consistency index.}
    \begin{algorithmic}[1]
    \Require The drug split result of 8 drug split scheme as known drug sets $D_k^i (i = 1,\ldots,8)$ and new drug sets $D_n^i(i = 1,\ldots,8)$. Threshold year for realistic drug split scheme as $y_t$. 
    \State Based on the threshold year $y_t$, split drugs with earlier approval time than $y_t$ into known drug set $D_k$ and drugs with later approval time than $y_t$ into new drug set $D_n$, which is the realistic drug split scheme.
    \For {$i = 1,\ldots,8$}
        \State $P_i = 0$.  // Initialize the penalty value
        \For {each drug $u \in D_k^i \cup D_n^i$}
            \If {$(u \in D_k^i \land u \in D_n) \lor (u \in D_n^i \land u \in D_k) $}
                \State $P_i = P_i +|y_u - y_t|$. 
            \EndIf
        \EndFor
    \EndFor
    \For {$i = 1,\ldots,8$}
        \State $C_i = \frac{\max_{i=1}^8 \{P_i\}}{P_i}$.
    \EndFor \\
    \Return Consistency index $C_i (i = 1,\ldots,8)$. 
    \end{algorithmic}
    \label{alg:consistency}
\end{algorithm}

\subsection{Significance Test for Consistency Index Comparison among Different Split Strategies}

\begin{figure*}[h]
    \centering
    \includegraphics[width=1.0\textwidth]{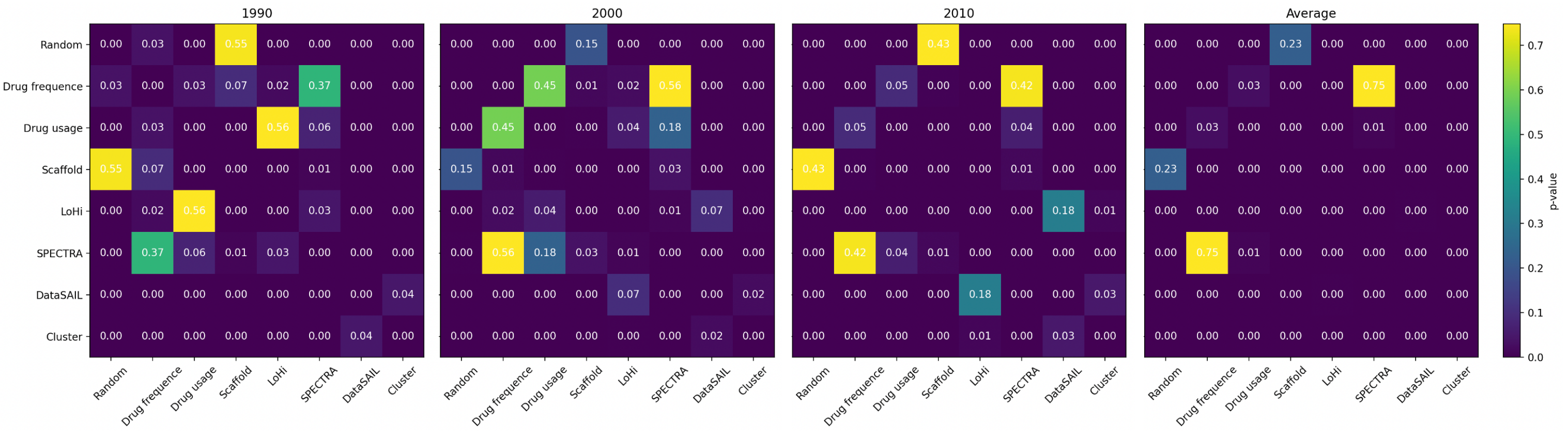}
    \caption{Significance test for consistency index comparison.}
    \label{fig:significance_test}
\end{figure*}

Figure~\ref{fig:significance_test} demonstrates the significance test for consistency index comparison among different data split strategies, which verifies that the proposed cluster-based split is more consistent with real-world data split. 

\subsection{Consistency Index Comparison among different similarity threshold $\gamma$ for cluster-based split}

Here we provide the consistency index with real-world drug split among different different similarity threshold $\gamma$ for cluster-based split. 
We can see that setting the normalized similarity threshold $\gamma$ as 1 could achieve highest consistency with realistic drug split scheme.  

\begin{figure}[t]
    \centering
    \includegraphics[width=0.4\textwidth]{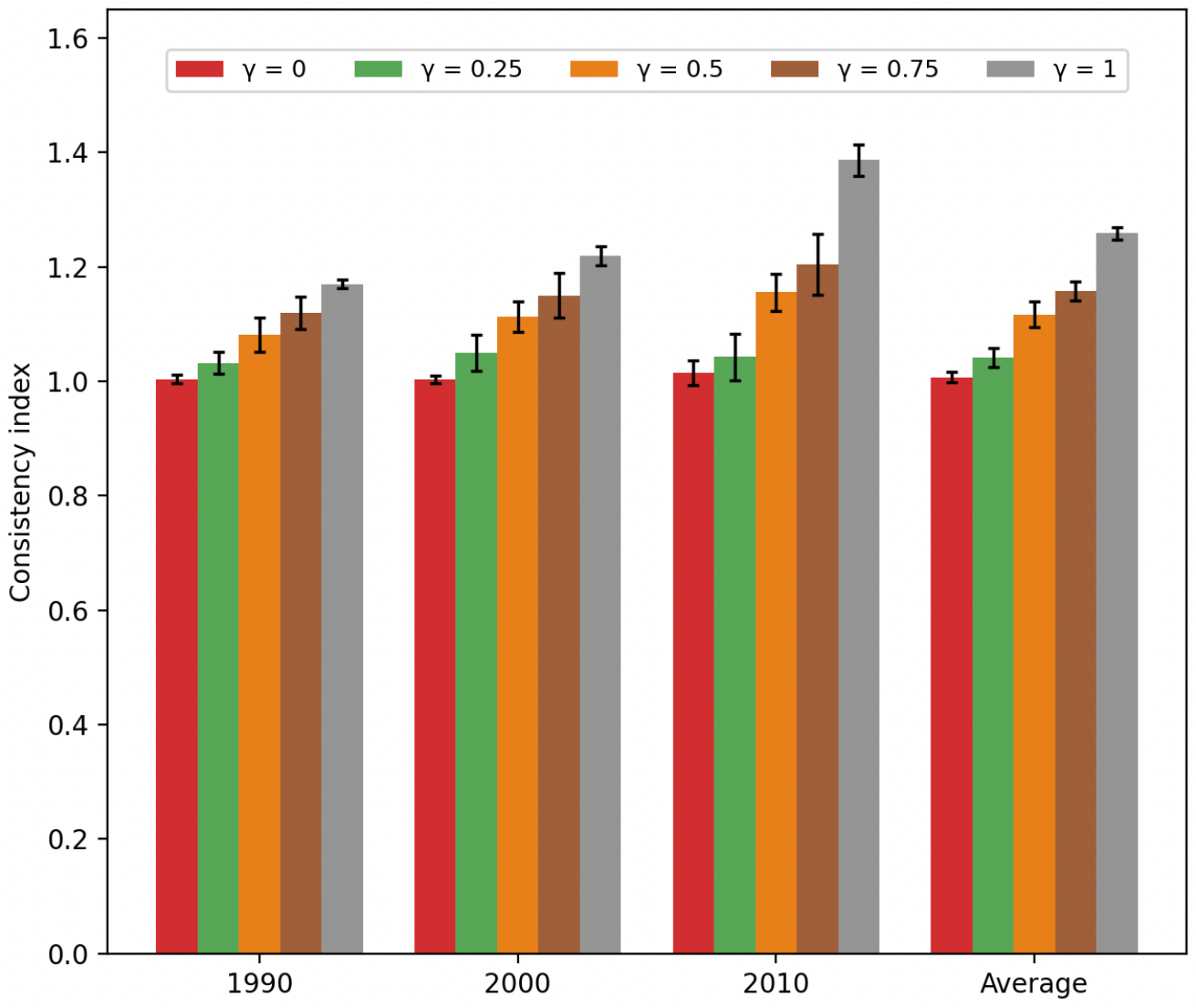}
	\vspace{-8px}
    \caption{Figure for time threshold of the realistic drug split scheme w.r.t the consistency index of different similarity threshold $\gamma$ for cluster-based split.}
    \label{fig:consistency2}
    \vspace{-15px}
\end{figure}

\subsection{Statistics of Datasets.}

Here we provide the general statistics of the two datasets used in experiments in Table~\ref{tab:dataset_count}.
Here $\mathcal{V}_\text{DDI}$ denotes the set of drugs, $\mathcal{R}_\text{DDI}$ denotes the set of drug-drug interaction types, and $\mathcal{N}_\text{DDI}$ denotes the set of DDI triplets.

\begin{table}[h]
    \caption{General statistics of two datasets.}
    \vspace{5px}
    \center
    \label{tab:dataset_count}
    \begin{tabular}{c|ccc}
        \toprule
        Dataset & $|\mathcal{V}_\text{DDI}|$ & $|\mathcal{R}_\text{DDI}|$ & $|\mathcal{N}_\text{DDI}|$ \\\midrule
        Drugbank & 1,710 & 86 & 188,509 \\
        TWOSIDES & 645 & 209 & 116,650 \\
        \midrule
    \end{tabular}
\end{table}

\subsection{Summary of Existing DDI methods.}
\label{app:ddi_methods}

Our summary of existing DDI prediction methods is shown in Table~\ref{tab:ddi_methods_part1}.

\begin{table*}[h]
    \caption{Summary of Drug-Drug Interaction Prediction Methods.}
    \small
    \renewcommand{\arraystretch}{0.80}
    \label{tab:ddi_methods_part1}
    \centering
    \begin{tabular}{>{\centering\arraybackslash}p{1.5cm}>{\centering\arraybackslash}p{2cm}>{\centering\arraybackslash}p{7.5cm}>{\centering\arraybackslash}p{5cm}}
        \toprule
        Method & Categorization & Description & Availability \\ 
        \midrule
        MLP & Feature based & Uses drug fingerprints to compute predicted logits. & $\times$ \\ 
        DeepDDI & Feature based & Takes drug chemical structures and names as inputs, predicting DDI types via DNN. & \url{https://bitbucket.org/kaistsystemsbiology/deepddi/src/master/} \\ 
        SFLLN & Feature based & Integrates four drug features and uses linear neighborhood regularization for DDIs.  & \url{https://bitbucket.org/kaistsystemsbiology/deepddi/src/master/} \\ 
        DDIMDL & Feature based & Constructs sub-models from drug features and combines them for DDI prediction.  & \url{https://github.com/BioMedicalBigDataMiningLabWhu/SFLLN.}{SFLLN} \\ 
        CSMDDI & Feature based & Uses RESCAL-based method to obtain drug and DDI type representations, learning a mapping function. &  \url{https://github.com/itsosy/csmddi} \\ 
        AMDE & Feature based & Uses graph and sequential encoders for drug features, with a multi-dimensional decoder for DDIs. &  \url{https://github.com/wan-Ying-Z/AMDE-master} \\
        \midrule
        ComplEx & Embedding based & Maps entities and relations into a complex vector space to calculate DDI probabilities. & $\times$ \\ 
        Graph-Embedding-4DDI & Embedding based & Applies RDF2Vec, TranE, and TransD to extract drug feature vectors. & \url{https://github.com/rcelebi/GraphEmbedding4DDI/} \\ 
        KG-DDI & Embedding based & Uses KG embedding methods and a Conv-LSTM network to predict DDI relations. & \url{https://github.com/rezacsedu/Drug-Drug-Interaction-Prediction} \\ 
        MSTE & Embedding based & Learns drug and relation embeddings, designing a score function for prediction. & \url{https://github.com/galaxysunwen/MSTE-master} \\ 
        DDKG & Embedding based & Learns global drug representations using neighboring embeddings and triple facts. & \url{https://github.com/Blair1213/DDKG} \\ 
        \midrule
        Decagon & GNN based & Uses drugs, genes, and diseases with GCN to update drug representations and predict DDIs. & \url{https://github.com/mims-harvard/decagon} \\ 
        SkipGNN & GNN based & Constructs a skip graph to obtain node embeddings, predicting DDIs via a decoder. & \url{https://github.com/kexinhuang12345/SkipGNN} \\ 
        KGNN & GNN based & Encodes drug and neighborhood info using GNN to predict DDIs. & \url{https://github.com/xzenglab/KGNN} \\ 
        SSI-DDI & GNN based & Models drug molecular graphs and predict DDI based on interaction between substructure of query drug pairs. & \url{https://github.com/kanz76/SSI-DDI} \\
        SumGNN & GNN based & Uses GNN to compute subgraph representations from an augmented network for DDI prediction. & \url{https://github.com/yueyu1030/SumGNN} \\ 
        DeepLGF & GNN based & Fuses local chemical structure, global, and biological function info for DDI prediction. & \url{https://github.com/MrPhil/DeepLGF} \\
        MRCGNN & GNN based &  Multi-relation graph contrastive learning strategy to better characteristics of rare DDI types. & \url{https://github.com/Zhankun-Xiong/MRCGNN} \\
        EmerGNN & GNN based & Uses flow based GNN with attention to update drug representations for DDIs. & \url{https://github.com/yzhangee/EmerGNN} \\ 
        KnowDDI & GNN based & Optimizes drug embeddings from augmented subgraphs to predict DDIs. & \url{https://github.com/LARS-research/KnowDDI} \\ 
        SAGAN & GNN based & Utilizes a transfer learning strategy to enhance the cross-domain generalization ability of GNNs. & \url{https://github.com/wyx2012/SAGAN}\\
        KSGTN-DDI & Graph-transformer based & Uses a Key Substructure-aware Graph Transformer for DDI prediction. & $\times$ \\ 
        TIGER & Graph-transformer based & Uses a Transformer based framework with self-attention and dual-channel network for DDI prediction. & \url{https://github.com/Blair1213/TIGER} \\ 
        DrugDAGT & Graph-transformer based & Uses dual-attention graph transformer with contrastive learning for DDI prediction. & \url{https://github.com/codejiajia/DrugDAGT} \\ 
        \midrule
        TextDDI & LLM based & Designs an LM-based predictor with RL-based selector for short DDI descriptions. & \url{https://github.com/zhufq00/DDIs-Prediction} \\ 
        DDI-GPT & LLM based & Uses knowledge graphs and pre-trained models to capture contextual dependencies for DDI prediction. & \url{https://github.com/Mew233/ddigpt} \\ 
        K-Paths & LLM based & Design a retrieval framework that extracts meaningful paths from KGs, enabling LLMs to predict unobserved drug-drug interactions. & \url{https://github.com/rsinghlab/K-Paths} \\ 
        \midrule
    \end{tabular}
\end{table*}

\subsection{Supplementary Information of Evaluation Metrics.} 
\label{app:metric}
According to evaluation metric mentioned in Section~\ref{sec:metric}, the evaluation metrics include F1-Score, accuracy and Cohen's Kappa for Drugbank: 
\begin{itemize}
\item F1-Score~(Macro) $ = \frac{1}{||\mathcal{P}_D||}\sum_{p\in\mathcal{P}_D}\frac{2P_p \cdot R_p}{P_p + R_p}$, where $P_p$ and $R_p$ are the precision and recall for the interaction type $p$, respectively. 
\item Accuracy: the proportion of correctly predicted interaction types relative to the ground-truth interaction types.
\item Cohen's Kappa: $\kappa = \frac{A_p - A_e}{1-A_e}$, where $A_p$ is the observed accuracy and $A_e$ is the probability of randomly seeing each class.
\end{itemize} 
And ROC-AUC, PR-AUC and accuracy for TWOSIDES:
\begin{itemize}
\item ROC-AUC $ = \sum_{k = 1}^n TP_k \Delta FP_k$ measures the area curve of receiver operating characteristics. 
$TP_k$ and $FP_k$ are the true positive rate and false positive rate at the $k$-th operating point.
\item PR-AUC $ = \sum_{k = 1}^n P_k \Delta R_k$ measures the area under curve of precision-recall. 
Here $P_k$ and $R_k$ are the precision and recall at the $k$-th operating point. 
\item Accuracy: the proportion of correctly predicted DDIs for each DDI type. 
\end{itemize}


\subsection{Model Training for Methods in DDI-Ben}
\label{app:training}
All the experiments in this work are conducted on a 24GB NVIDIA GeForce RTX 4090 GPU for 200 epochs. 
To obtain the best performance of existing computational DDI methods, we conduct comprehensive hyper-parameter tuning for each method. 
Table~\ref{tab:hyp} shows a comprehensive list of hyper-parameters for each method, where we reference the tunning space of the original works.  
Hyperparameter tunning is conducted via Bayesian optimization method. 

\begin{table}[h]
    \caption{Hyperparameter search space for all compared methods.}
    \small
    \center
    \setlength\tabcolsep{2pt}
    \label{tab:hyp}
    \begin{tabular}{c|l|c}
        \toprule
        Method & Hyperparameter & Search Space \\ \midrule
        General  & Learning rate & [0.0001, 0.0003, 0.001, 0.003] \\
        Settings& Weight decay & [1e-6, 1e-5, 1e-4, 0] \\ 
        & Dropout rate & [0, 0.1, 0.2, 0.3, 0.4, 0.5] \\ 
        & Batch size & [64, 128, 256] \\
        & Training epoch & [100] \\ \midrule
        MLP &Layer number & [1, 2, 3] \\
        & Hidden dimension& [50, 100, 200] \\ \midrule
        MSTE &Embedding dimension & [50, 100, 200] \\ \midrule
        Decagon &Layer number& [1, 2, 3] \\
        & Hidden dimension& [50, 100, 200] \\ \midrule
        SSI-DDI & Hidden dimension& [32, 64] \\ \midrule
        MRCGNN & Hidden dimension& [32, 64] \\ \midrule
        EmerGNN & Subgraph sampling hop & [1, 2, 3, 4] \\
        &Hidden dimension& [32, 64] \\\midrule
        SAGAN & Hidden dimension& [32, 64] \\ \midrule
        TIGER &Layer number& [1, 2, 3] \\
        &Hidden dimension& [32, 64] \\
        \midrule
    \end{tabular}
\end{table}

\subsection{The Meaning of Selected DDI Types in Section~\ref{sec:relation} in Drugbank Dataset}

\begin{itemize}
\item \#48: The risk or severity of adverse effects can be increased when \#Drug1 is combined with \#Drug2.
\item \#46: The metabolism of \#Drug2 can be decreased when combined with \#Drug1. 
\item \#72: The serum concentration of \#Drug2 can be increased when it is combined with \#Drug1.
\item \#29: \#Drug1 may increase the orthostatic hypotensive activities of \#Drug2. 
\item \#71: \#Drug1 may decrease the excretion rate of \#Drug2 which could result in a higher serum level.
\item \#57: \#Drug1 may decrease the cardiotoxic activities of \#Drug2.
\item \#24: \#Drug1 may increase the atrioventricular blocking~(AV block) activities of \#Drug2.
\item \#1: \#Drug1 may increase the photosensitizing activities of \#Drug2.
\item \#18: \#Drug1 may increase the vasoconstricting activities of \#Drug2.
\end{itemize}

\subsection{Construction of Real-world Emerging DDI Prediction Dataset}
\label{app:approval_time}

\begin{figure*}[h]
    \centering
    \includegraphics[width=0.8\textwidth]{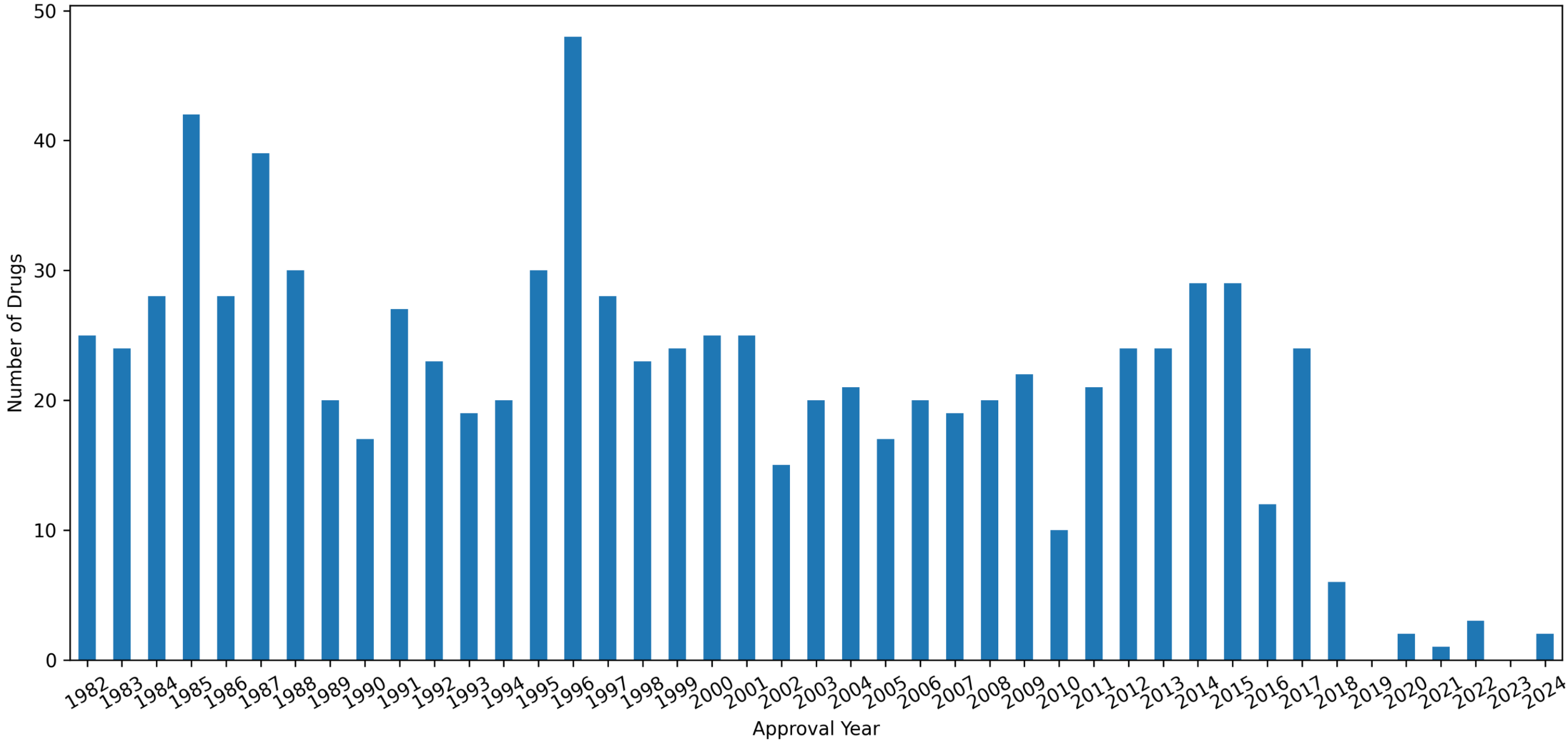}
    \caption{The distribution of available approval time of drugs in Drugbank dataset.}
    \label{fig:approval_time}
\end{figure*}

In this work, we use the Drugbank dataset to construct the real-world emerging DDI prediction dataset. 
We first extract 1710 drugs and their names from the DDI dataset. 
Then we collect the approval times of these drugs based on their names from FDA Drugs Database~\url{https://www.drugfuture.com/fda/}. 
Among these drugs, totally 886 drugs have available approval time information, with their distribution shown in Figure~\ref{fig:approval_time}.

\subsection{Limitations of DDI-Ben}
\label{app:limitations}

In DDI-Ben, we emphasize the importance of distribution changes that could greatly affect the performance of emerging DDI prediction performance in realistic drug development scenarios. 
Although we provide an ensembling method of best-performing existing DDI prediction techniques to handle the negative impact of distribution changes, its incremental gains remain notably limited.
New methods that can effectively improve the performance of DDI prediction under distribution changes are still needed, which is also the future work of this paper.

\clearpage

\section{Additional Experimental Results}
\label{app:add}

\subsection{Additional Experimental Results for General Method Performance}

\begin{figure}[h]
	\centering
	
	\vspace{-7px}
	\includegraphics[width=0.7\textwidth]{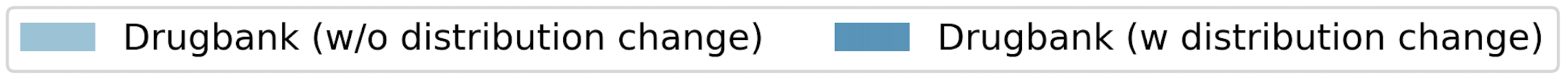}
    \vspace{-8px}
    
	\subfigure[\normalsize Accuracy in S1 task on Drugbank. ]
	{\includegraphics[width=0.48\textwidth]{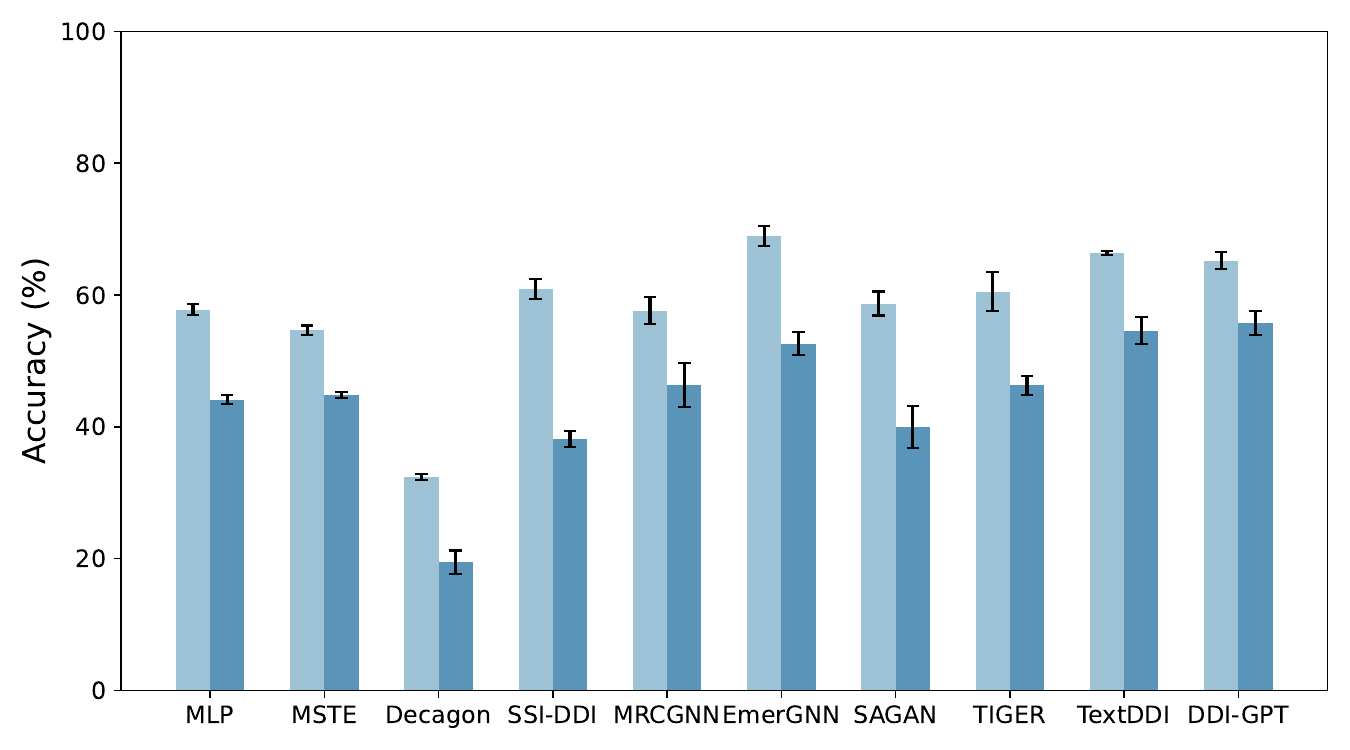}}
	\subfigure[\normalsize \normalsize Accuracy in S2 task on Drugbank. ]
	{\includegraphics[width=0.48\textwidth]{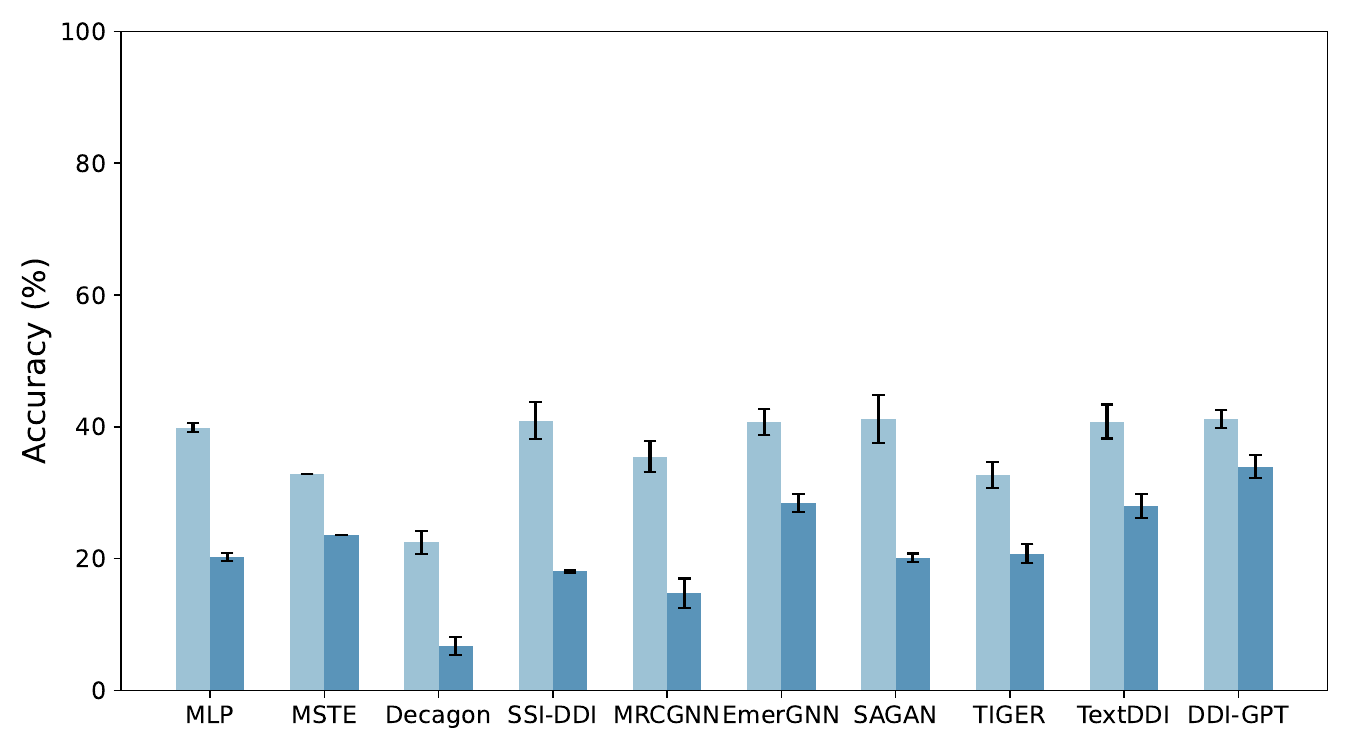}}
    \vspace{-5px}
    
	\subfigure[\normalsize Cohen's Kappa in S1 task on Drugbank. ]
	{\includegraphics[width=0.48\textwidth]{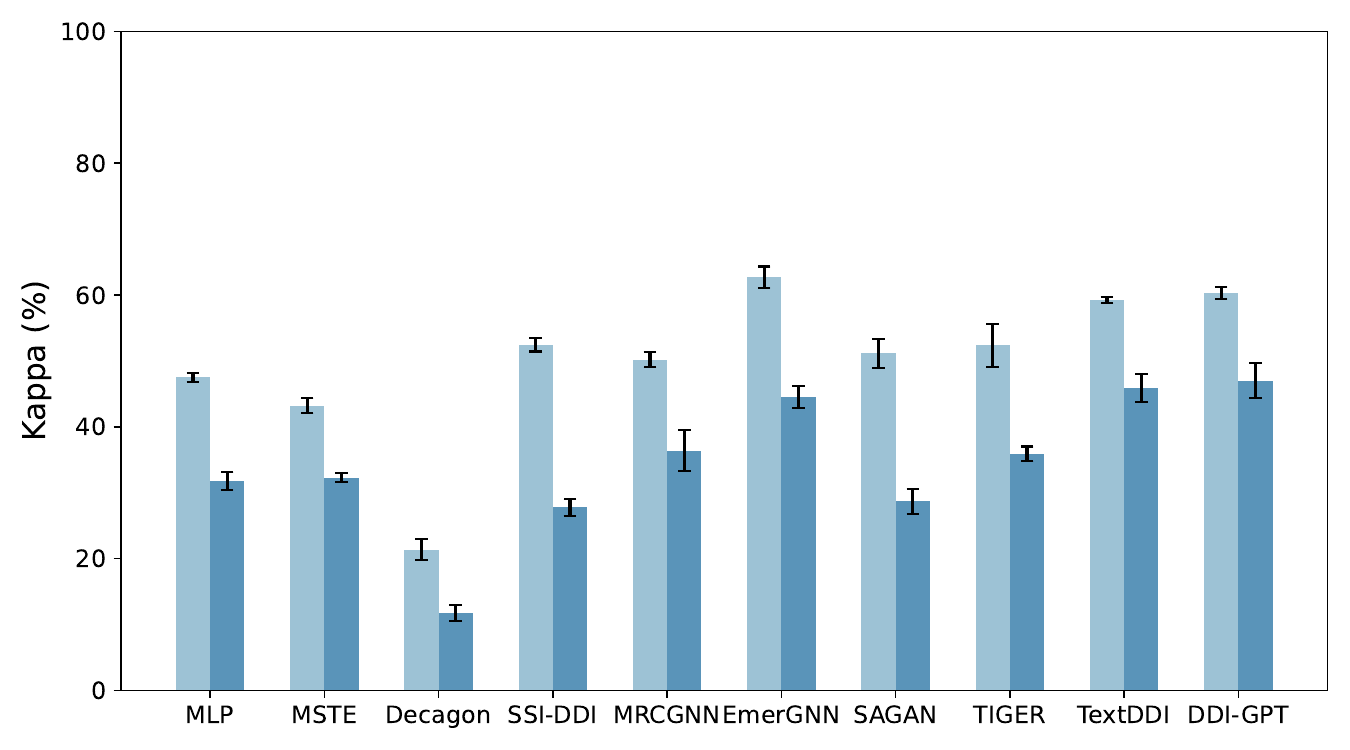}}
	\subfigure[\normalsize Cohen's Kappa in S2 task on Drugbank. ]
	{\includegraphics[width=0.48\textwidth]{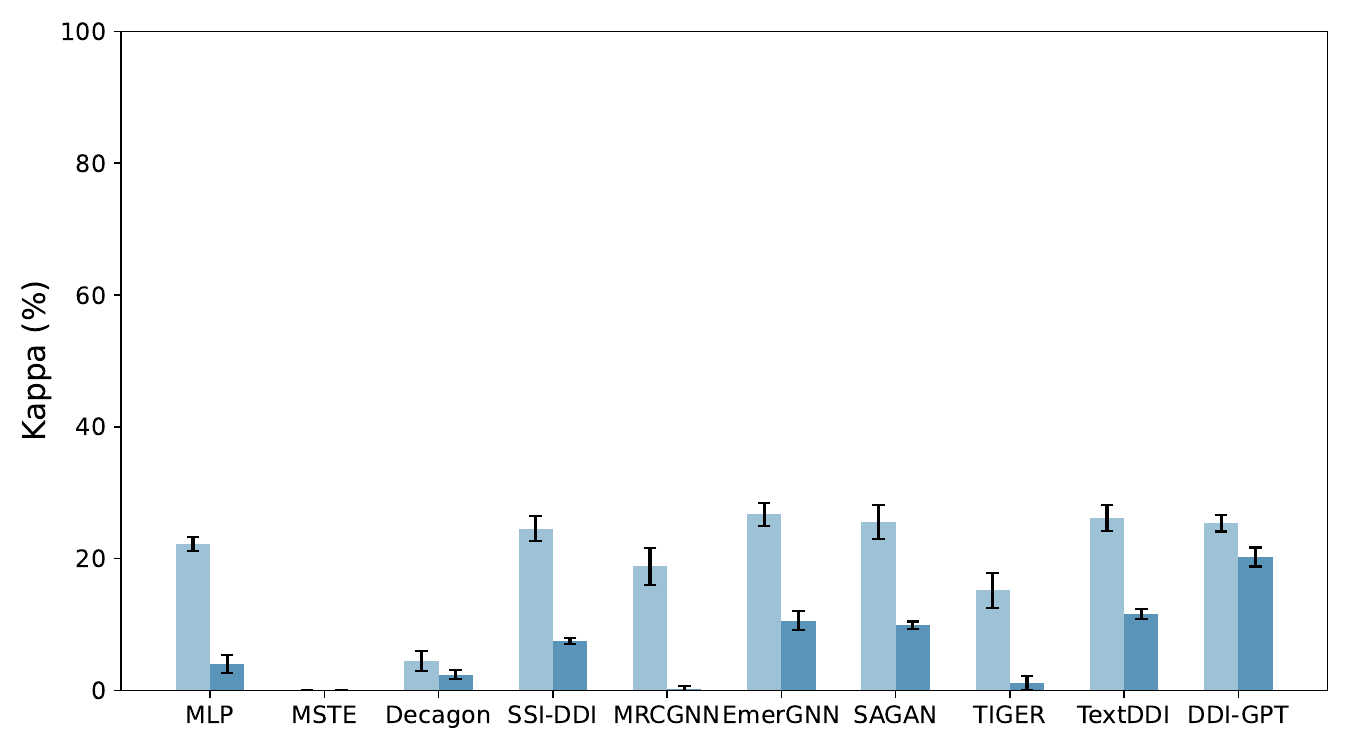}}

    \vspace{5px}

	\caption{Performance comparison in the setting with and without distribution change in S1-S2 tasks. 
	Here is the results that use accuracy and Cohen's Kappa as evaluation metrics for Drugbank. }
    \label{fig:overview_add1}
    \vspace{-10px}
\end{figure}

\begin{figure}[h]
	\centering
	
	{\includegraphics[width=0.7\textwidth]{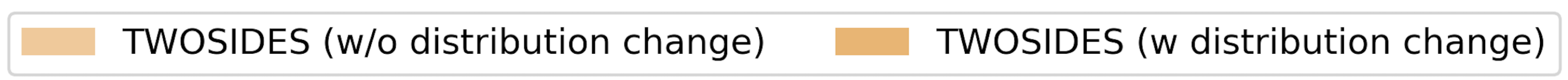}}
    \vspace{-8px}
	\subfigure[\normalsize Accuracy in S1 task on TWOSIDES. ]
	{\includegraphics[width=0.48\textwidth]{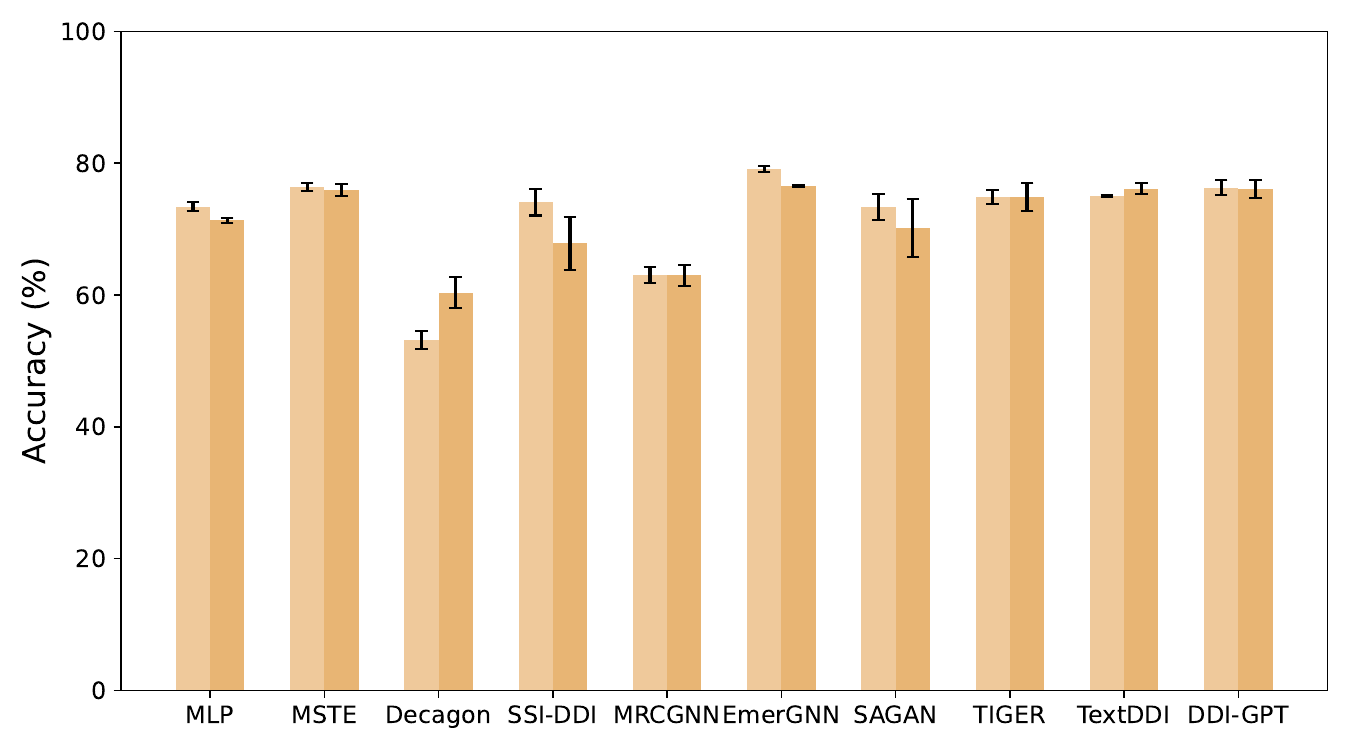}}
	\subfigure[\normalsize Accuracy in S2 task on TWOSIDES. ]
	{\includegraphics[width=0.48\textwidth]{fig_app/TWOSIDES_Accuracy_S1_comp1.pdf}}
    \vspace{-8px}
	\subfigure[\normalsize PR-AUC in S1 task on TWOSIDES. ]
	{\includegraphics[width=0.48\textwidth]{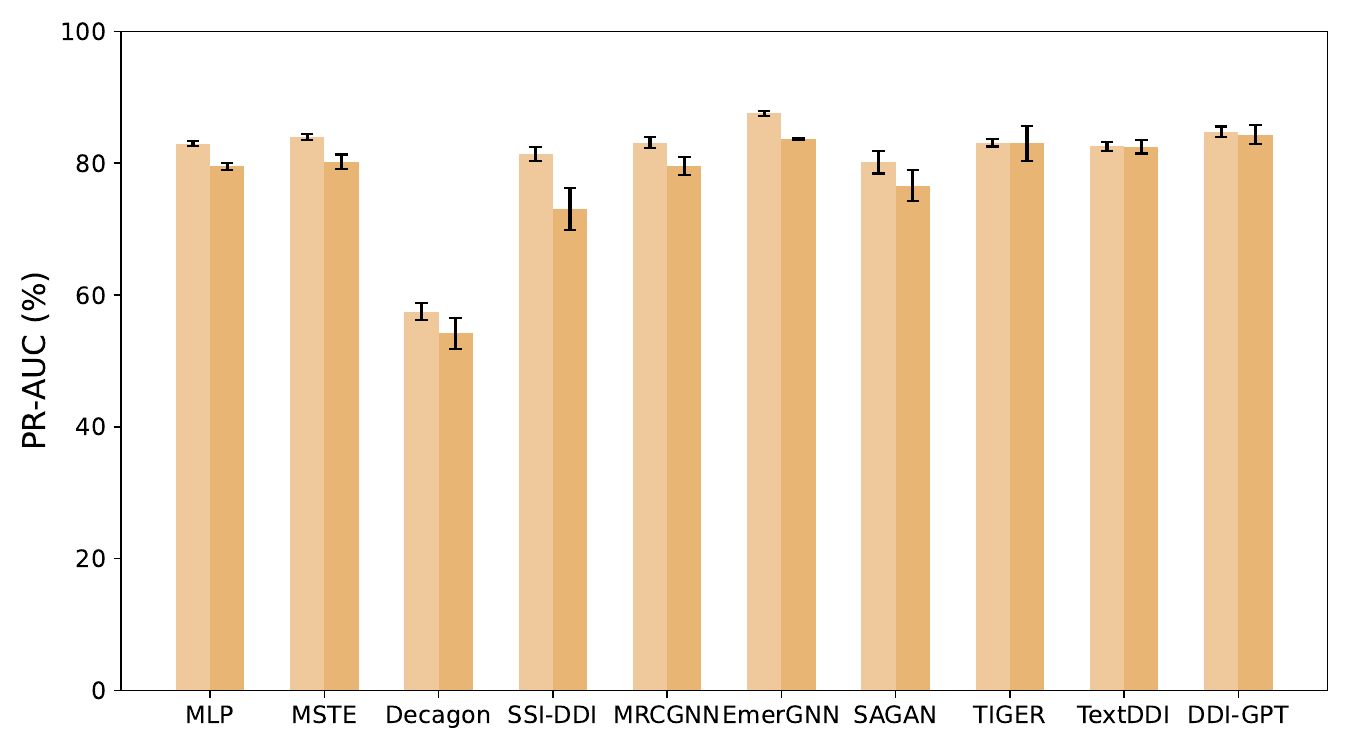}}
	\subfigure[\normalsize PR-AUC in S2 task on TWOSIDES. ]
	{\includegraphics[width=0.48\textwidth]{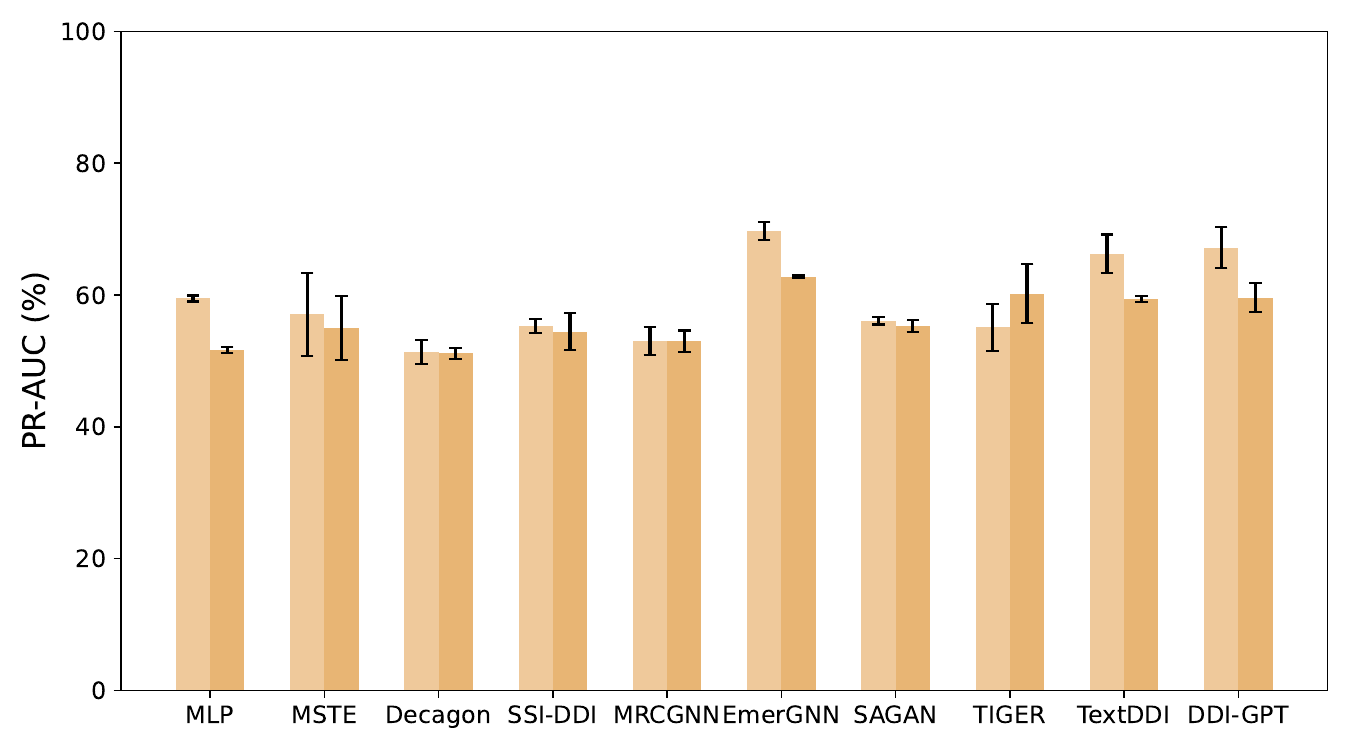}}
    \vspace{5px}

	\caption{Performance comparison in the setting with and without distribution change in S1-S2 tasks. 
	Here is the results that use PR-AUC and accuracy as evaluation metrics for TWOSIDES. }
    \label{fig:overview_add2}
    \vspace{-10px}
\end{figure}

\clearpage

\subsection{Relative Performance Comparison of Different Methods on Drugbank and TWOSIDES Datasets}
\label{app:rela}

\begin{table}[H]
    \caption{Comparison of different methods on Drugbank dataset \textbf{without} and \textbf{with} distribution change. “Emb” is short for “Embedding”; “GT” is short for “Graph-transformer”. The evaluation metrics are presented in percentage (\%) with best values in boldface. Avg. Score is the mean of the scores across two settings (S1 and S2) for each method, representing overall performance. Rank is determined based on the Avg. Score, with the highest Avg. Score receiving rank 1.}
    \vspace{5px}
    \centering
    \label{tab:comparison_on_Drugbank}
    \resizebox{\textwidth}{!}{
        \begin{tabular}{cc|cccc|cccc}
            \toprule
            \multicolumn{2}{c|}{Datasets} & \multicolumn{4}{c|}{\textbf{Drugbank (F1-Score, w/o distribution change)}} & \multicolumn{4}{c}{\textbf{Drugbank (F1-Score, w distribution change)}} \\ 
            \midrule
            Category & Methods & S1 & S2 & Avg. Score & Rank & S1 & S2 & Avg. Score & Rank \\
            \midrule
            Feature & MLP &  12.5\textpm2.3 &  9.0\textpm1.9 & 10.8 & 8 &  9.0\textpm2.1 &  2.8\textpm0.6 & 5.9 & 9 \\
            \midrule
            Emb & MSTE &  14.5\textpm0.7 &  1.0\textpm0.0 & 7.8 & 9 &  12.3\textpm0.3 &  1.2\textpm0.0 & 6.8 & 8  \\
            \midrule
            GNN & Decagon &  11.6\textpm0.5 &  2.9\textpm0.0 & 7.3 & 10 &  4.1\textpm0.0 &  1.3\textpm0.1 & 2.7 & 10  \\
            & SSI-DDI &  50.5\textpm1.8 &  19.1\textpm1.3 & 34.8 & 5 &  19.1\textpm0.1 &  6.5\textpm0.5 & 12.8 & 6  \\
            & MRCGNN &  41.7\textpm1.3 &  8.6\textpm1.6 & 25.2 & 7 &  12.0\textpm2.8 &  2.3\textpm0.9 & 7.1 & 7  \\
            & EmerGNN &  56.9\textpm1.7 & \textbf{ 22.5\textpm1.2} & \textbf{39.7} & 1 &  34.0\textpm2.1 &  3.1\textpm0.3 & 18.6 & 3  \\
            & SAGAN &  51.3\textpm0.6 &  18.8\textpm2.2 & 35.0 & 4 &  22.3\textpm2.6 &  8.1\textpm1.3 & 15.2 & 4  \\
            \midrule
            GT & TIGER &  47.0\textpm2.5 &  11.9\textpm2.0 & 29.5 & 6 &  25.7\textpm1.4 &  2.9\textpm1.2 & 14.3 & 5  \\
            \midrule
            LLM & TextDDI &  56.8\textpm0.8 &  18.2\textpm0.2 & 37.5 & 3 & \textbf{ 36.7\textpm1.0} &  10.9\textpm1.0 & 23.8 & 2  \\
            & DDI-GPT &  \textbf{57.3\textpm1.4} &  18.8\textpm2.2 & 38.0 & 2 &  36.4\textpm1.9 &  \textbf{11.6\textpm1.8} & \textbf{24.0} & 1  \\
            \midrule
        \end{tabular}
    }
\end{table}

\begin{table}[H]
    \caption{Comparison of different methods on TWOSIDES dataset \textbf{without} and \textbf{with} distribution change.}
    \vspace{5px}
    \centering
    \label{tab:comparison_on_TWOSIDES}
    \resizebox{\textwidth}{!}{
        \begin{tabular}{cc|cccc|cccc}
            \toprule
            \multicolumn{2}{c|}{Datasets} & \multicolumn{4}{c|}{\textbf{TWOSIDES (ROC-AUC, w/o distribution change)}} & \multicolumn{4}{c}{\textbf{TWOSIDES (ROC-AUC, w distribution change)}}  \\ 
            \midrule
            Category & Methods & S1 & S2 & Avg. Score & Rank & S1 & S2 & Avg. Score & Rank \\
            \midrule
            Feature & MLP &  84.7\textpm0.2 &  60.0\textpm0.6 & 72.4 & 4 &  78.8\textpm0.5 &  47.1\textpm0.6 & 63.0 & 9  \\
            \midrule
            Emb& MSTE &  86.1\textpm0.3 &  57.0\textpm6.5 & 71.4 & 5 &  82.4\textpm1.0 &  52.7\textpm7.0 & 67.6 & 5  \\
            \midrule
            GNN & Decagon &  59.3\textpm1.7 &  49.5\textpm2.8 & 54.4 & 10 &  57.5\textpm2.1 &  50.8\textpm0.8 & 54.2 & 10  \\
            & SSI-DDI &  81.5\textpm1.3 &  55.6\textpm0.9 & 68.5 & 7 &  74.4\textpm3.8 &  54.7\textpm2.4 & 64.5 & 8  \\
            & MRCGNN &  83.4\textpm0.4 &  53.2\textpm1.9 & 68.3 & 8 &  79.8\textpm1.1 &  53.4\textpm1.4 & 66.6 & 6  \\
            & EmerGNN &  \textbf{86.4\textpm0.5} & \textbf{ 72.5\textpm1.9} & \textbf{79.5} & 1 &  84.8\textpm0.2 & \textbf{ 60.5\textpm0.3} & 72.7 & 2  \\
            & SAGAN &  80.9\textpm0.8 &  56.4\textpm0.8 & 69.2 & 6 &  77.0\textpm3.0 &  55.6\textpm0.7 & 66.3 & 7  \\
            \midrule
            GT & TIGER &  82.7\textpm0.5 &  52.5\textpm2.9 & 67.6 & 9  &  84.6\textpm1.7 &  58.7\textpm5.0 & 71.7 & 4  \\
            \midrule
            LLM & TextDDI &  83.3\textpm0.6 &  68.7\textpm1.2 & 76.0 & 3 & \textbf{ 85.5\textpm0.7} &  60.3\textpm2.3 & \textbf{72.9} & 1  \\
            & DDI-GPT &  85.2\textpm0.7 &  68.5\textpm2.8 & 76.8 & 2 &  85.0\textpm1.6 &  60.4\textpm1.9 & 72.7 & 2  \\
            \midrule
        \end{tabular}
    }
\end{table}


\clearpage

\subsection{Additional Results for Experiment on DDI Types}

The results of DDI prediction performance for different DDI types on Drugbank in S2 task are shown in Table~\ref{tab:type2}.

\begin{table}[H]
	\caption{DDI prediction performance for different DDI types on Drugbank~(S2 task). Here ``Major'', ``Medium'', ``Long-tail'' denote DDI types with high, medium, low occurrence frequency, respectively.
    ``w/o'' and ``w'' denote the setting without and with distribution change introduced.
    For each DDI type, the best results for ``w/o'' and ``w'' setting are marked by \underline{underline} and \textbf{bold}, respectively.
    }
	\footnotesize
	\label{tab:type2}
    \setlength\tabcolsep{8pt}
	\begin{center}
		\begin{tabular}{cc|ccc|ccc|ccc}
			\toprule
           \multirow{2}{*}{Method} & Distribution & \multicolumn{3}{c|}{Major} & \multicolumn{3}{c|}{Medium} & \multicolumn{3}{c}{Long-tail}\\
            & change & \#48 & \#46 & \#72 & \#29 & \#71 & \#57 & \#24 & \#1 & \#18 \\ \midrule
            \multirow{2}{*}{MLP} &w/o & 74.5 & 22.9 & \underline{41.8} & 16.7 & 59.3 & 22.2 & 11.1 & 28.6 & 16.7 \\
             &w & 58.4 & 13.9 & 20.2 & 0.0 & 0.0 & 7.1 & 0.0 & 0.0 & 0.0 \\ \midrule
            \multirow{2}{*}{MSTE} &w/o & \underline{100.0} & 0.0 & 0.0 & 0.0 & 0.0 & 0.0 & 0.0 & 0.0 & 0.0 \\ 
             &w & \textbf{100.0} & 0.0 & 0.0 & 0.0 & 0.0 & 0.0 & 0.0 & 0.0 & 0.0 \\ \midrule
            \multirow{2}{*}{EmerGNN} &w/o & 67.0 & \underline{38.4} & 34.1 & \underline{37.5} & 85.2 & \underline{43.3} & \underline{37.8} & \underline{82.9} & 0.0 \\
             &w & 74.0 & \textbf{41.2} & 8.0 & 8.0 & 0.0 & 0.0 & 0.0 & 0.0 & 0.0 \\ \midrule
            \multirow{2}{*}{TIGER} &w/o & 65.5 & 16.7 & 22.4 & 16.7 & \underline{93.0} & 22.2 & 7.4 & 28.6 & 50.0 \\
             &w & 61.5 & 2.3 & 1.3 & 5.0 & 7.4 & 2.4 & 0.0 & 5.6 & 16.7 \\ \midrule
            \multirow{2}{*}{DDI-GPT} &w/o & 80.4 & 19.1 & 37.1 & 30.6 & 68.2 & 38.1 & 30.4 & 58.0 & \underline{66.7} \\
             &w & 66.2 & 13.7 & \textbf{23.6} & \textbf{27.7} & \textbf{24.7} & \textbf{43.6} & \textbf{17.4} & \textbf{26.3} & \textbf{41.7} \\ \midrule
		\end{tabular}
	\end{center}
\end{table}

\clearpage

\begin{figure}[H]
	\centering
	
	\subfigure[\normalsize Accuracy w/o distribution change in S1 task. ]
	{\includegraphics[width=0.45\textwidth]{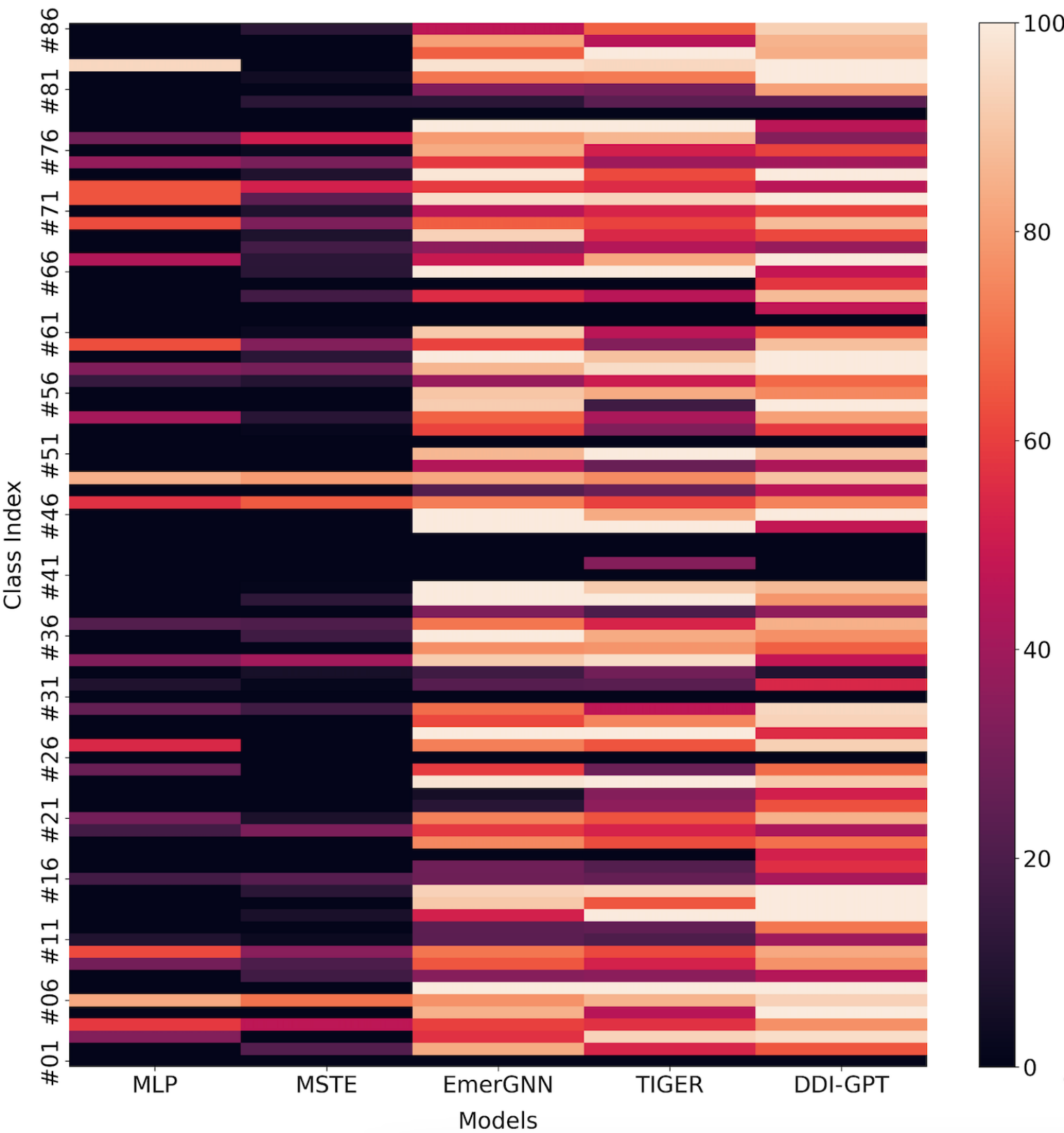}}
	\subfigure[\normalsize Accuracy w distribution change in S1 task. ]
	{\includegraphics[width=0.52\textwidth]{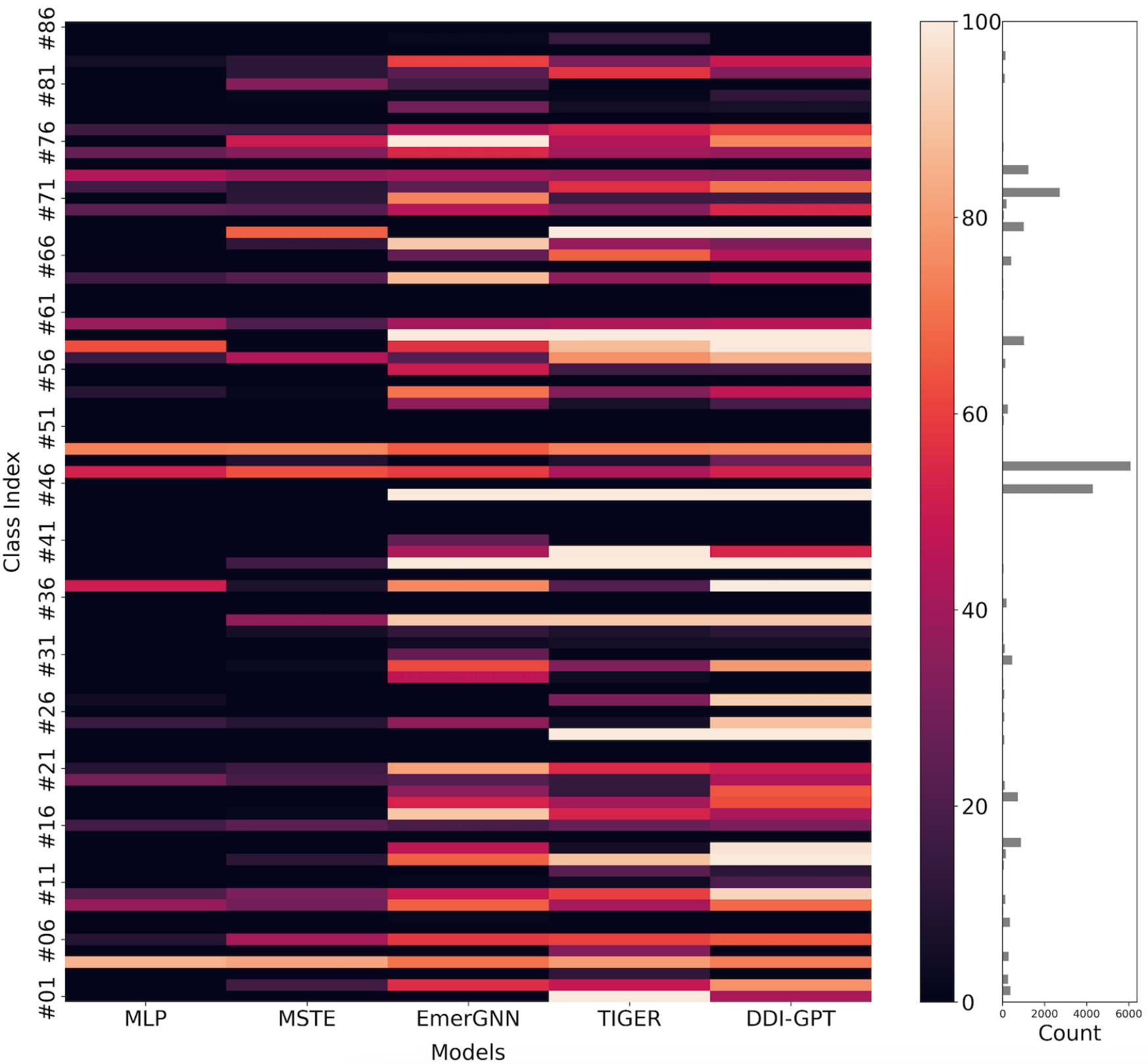}}
	\subfigure[\normalsize Accuracy w/o distribution change in S2 task. ]
	{\includegraphics[width=0.45\textwidth]{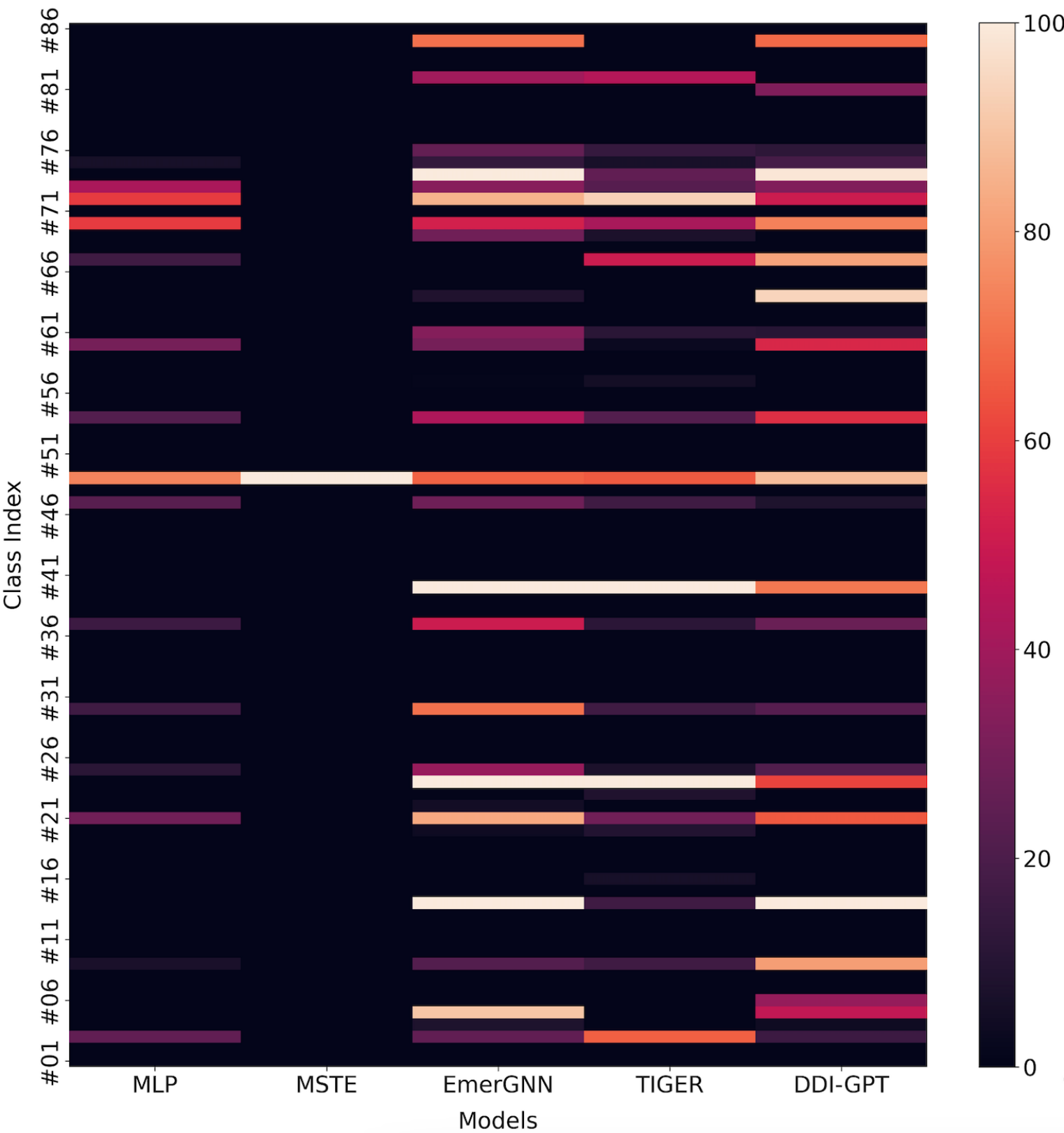}}
	\subfigure[\normalsize Accuracy w distribution change in S2 task. ]
	{\includegraphics[width=0.52\textwidth]{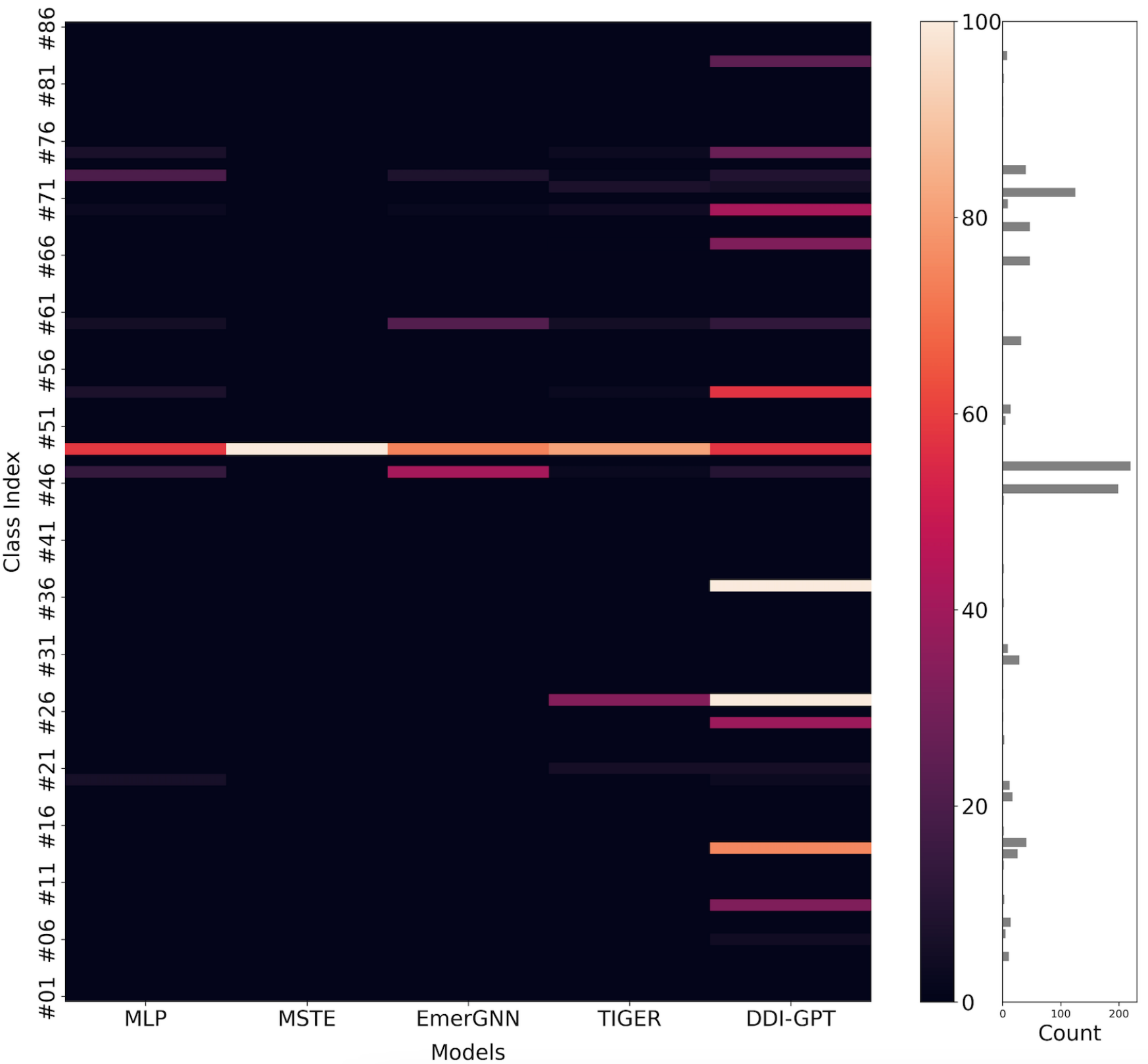}}
	\caption{The performance of representative DDI prediction methods for all DDI types on Drugbank dataset. }
	\label{fig:all_type}
\end{figure}

The performance of representative DDI prediction methods on all DDI types in Drugbank dataset is shown in Figure~\ref{fig:all_type}. 
Here the frequency of different DDI types is shown on the right.

\clearpage

\subsection{Additional Experimental Results for Controlling Distribution Change}

\begin{figure}[H]
	\centering
	\includegraphics[width=0.5\textwidth]{fig/different_gamma_legend.pdf}
    \vspace{-10px}
	
	\subfigure[\normalsize Accuracy for S1 task on Drugbank. ]
	{\includegraphics[width=0.45\textwidth]{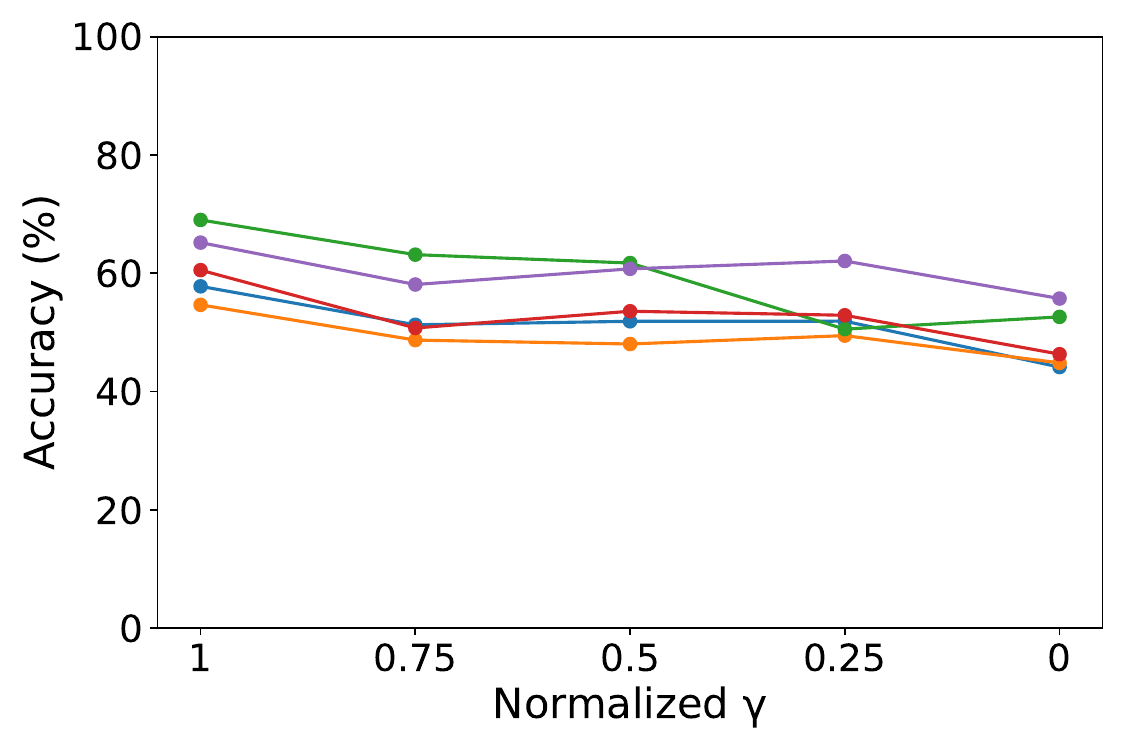}}
	\subfigure[\normalsize Accuracy for S2 task on Drugbank. ]
	{\includegraphics[width=0.45\textwidth]{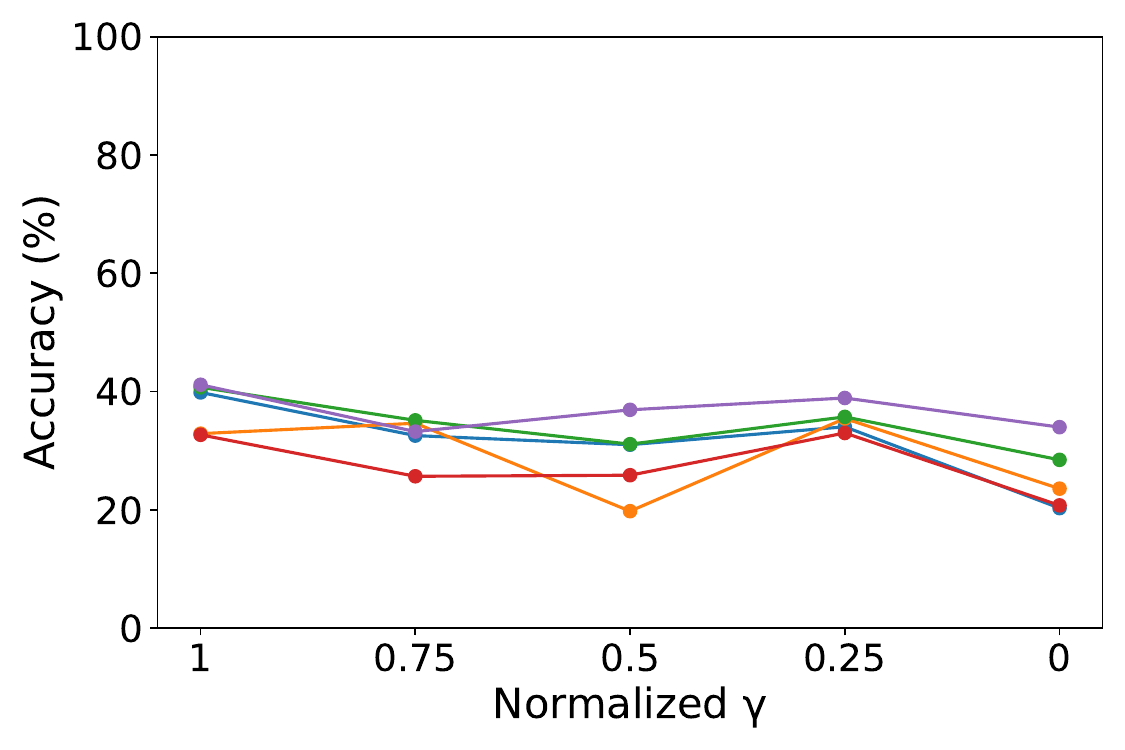}}
	\subfigure[\normalsize Cohen's Kappa for S1 task on Drugbank. ]
	{\includegraphics[width=0.45\textwidth]{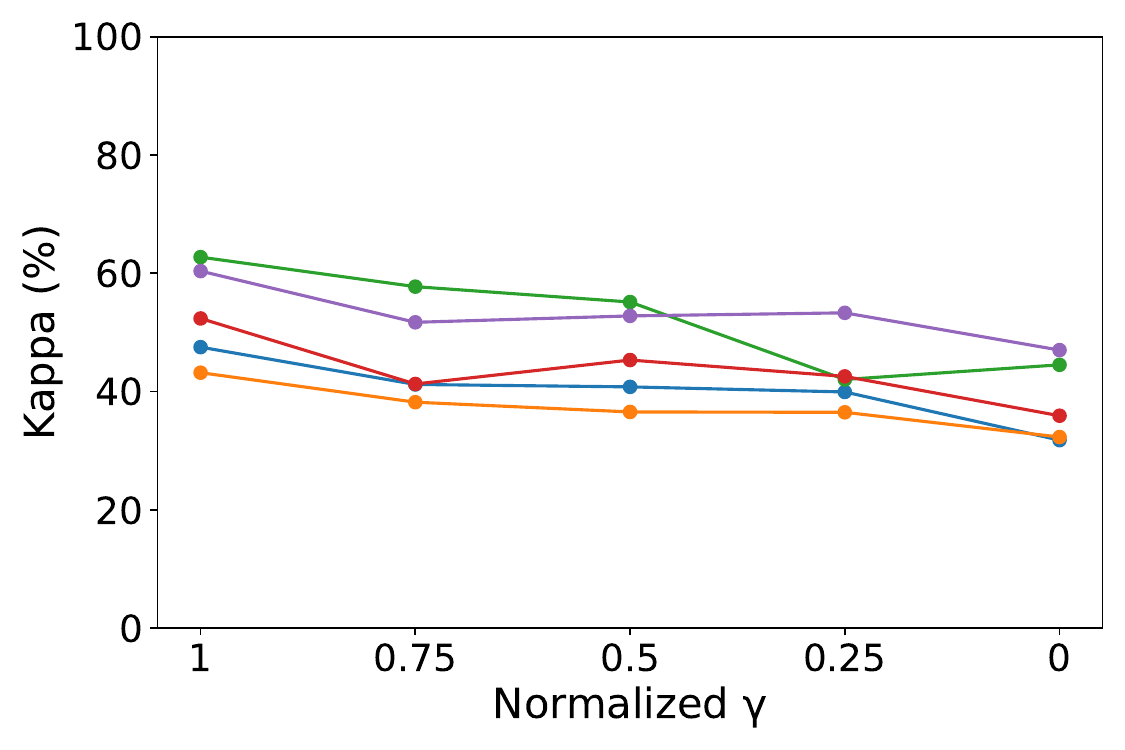}}
	\subfigure[\normalsize Cohen's Kappa for S2 task on Drugbank. ]
	{\includegraphics[width=0.45\textwidth]{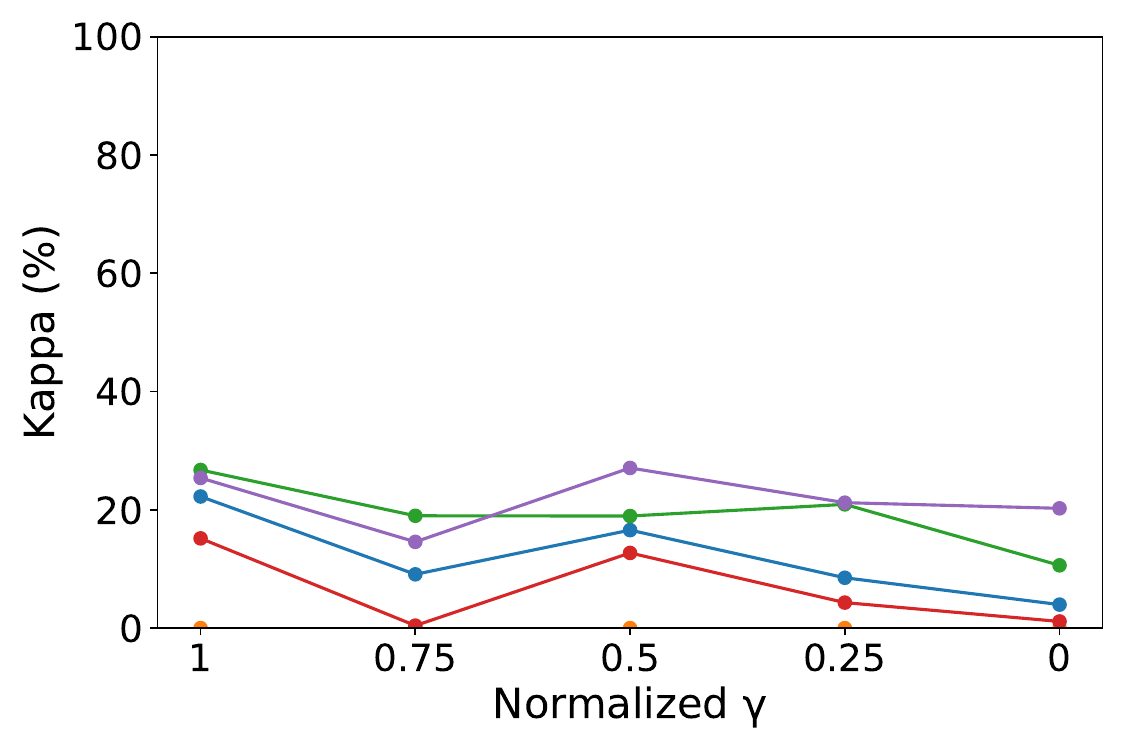}}
	\caption{Tunning $\gamma$ in the setting with distribution change on Drugbank dataset~(accuracy and Cohen's Kappa as evaluation metric). }
	\label{fig:gamma_f1f}
	\vspace{-8px}
\end{figure}

\clearpage

\subsection{Benchmarking Results Using DataSAIL to Conduct Drug Split}
\label{app:sail}

\begin{figure*}[h]
	\centering
	\includegraphics[width=0.6\textwidth]{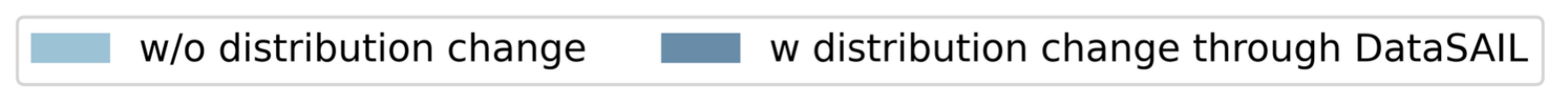}
	\vspace{-5px}
	
	\subfigure[\normalsize S1 task on Drugbank. ]
	{\includegraphics[width=0.48\textwidth]{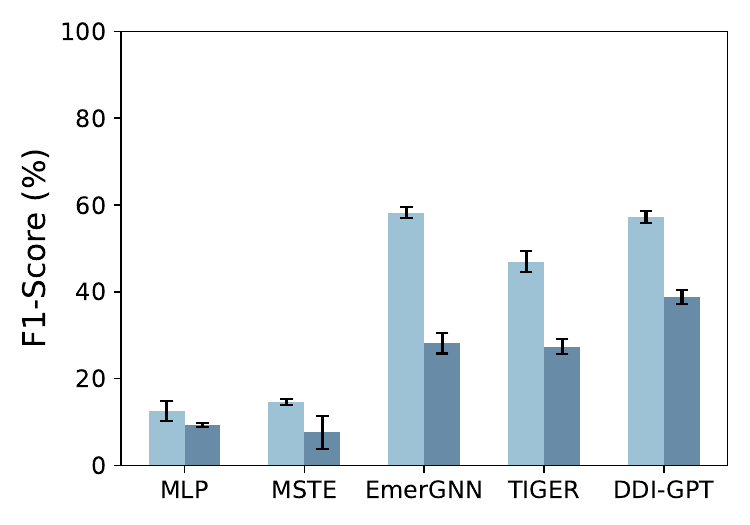}}
    \qquad
	\subfigure[\normalsize S2 task on Drugbank. ]
	{\includegraphics[width=0.48\textwidth]{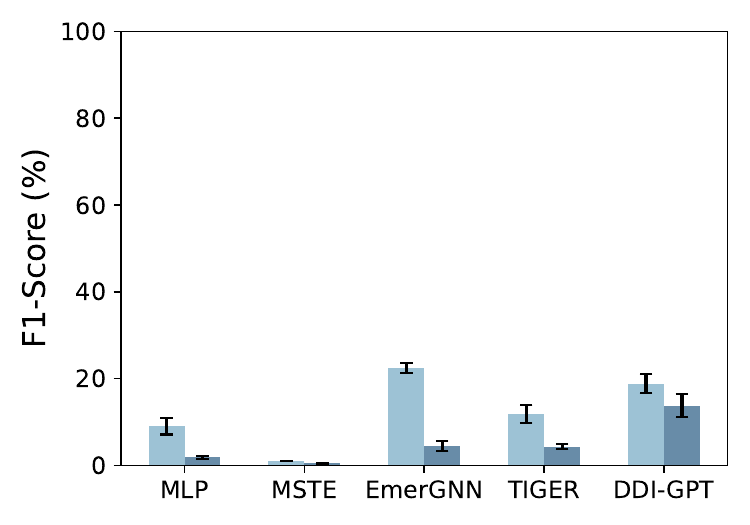}}
	
	\includegraphics[width=0.6\textwidth]{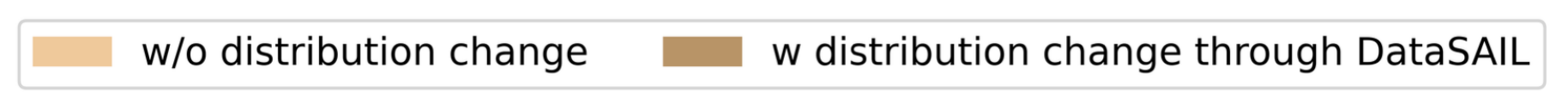}
	\vspace{-5px}
	
	\subfigure[\normalsize S1 task on TWOSIDES. ]
	{\includegraphics[width=0.48\textwidth]{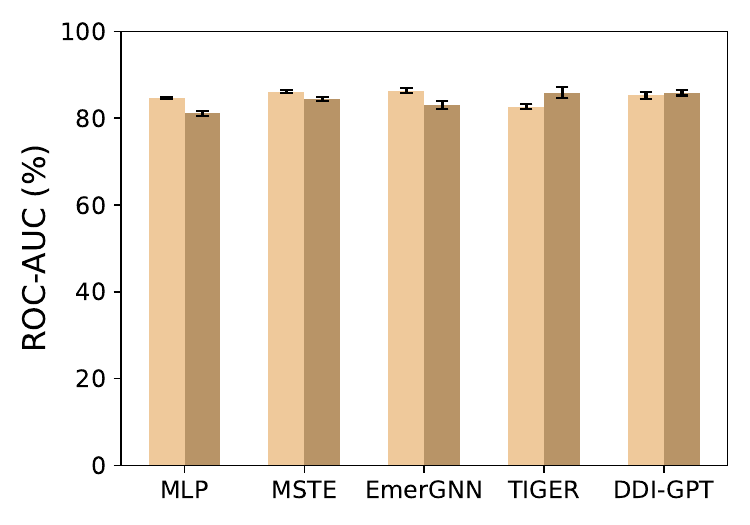}}
    \qquad
	\subfigure[\normalsize S2 task on TWOSIDES. ]
	{\includegraphics[width=0.48\textwidth]{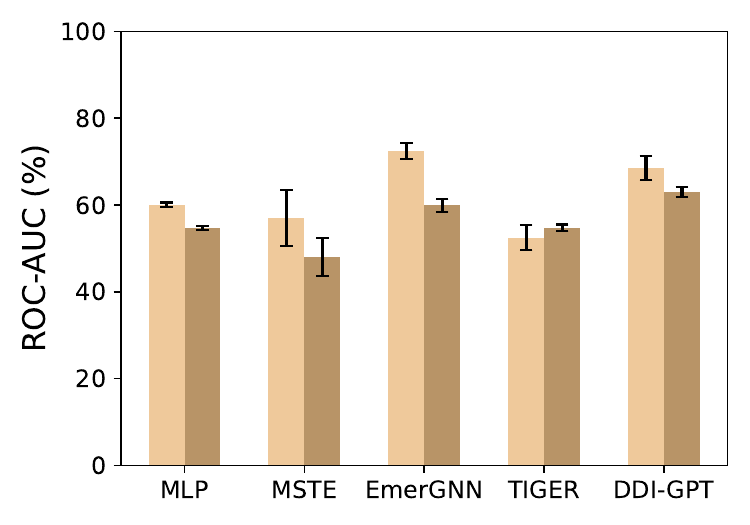}}
	\caption{Performance comparison for different types of DDI methods in the settings with and without distribution change in S1-S2 tasks. 
    Here drug splitting for known and new drugs is conducted by DataSAIL and we utilize primary evaluation metric for each dataset~
		(F1 for Drugbank and ROC-AUC for TWOSIDES).}
	\label{fig:overview_sail}
	\vspace{-5px}
\end{figure*}

In consistency comparison among different drug split scheme in Section~\ref{sec:consistency}, we can see that DataSAIL achieves the second highest consistency index with realistic drug split scheme. 
Here we further conduct experiments using DataSAIL to conduct distribution change simulation and compare the results with the setting without distribution change.
As shown in Figure~\ref{fig:overview_sail}, we can see that GNN based method~(EmerGNN), graph transformer based method~(TIGER) and LLM based methods~(DDI-GPT) still achieve relatively better performance.
The performance degradation of various methods under distribution changes remains substantial compared with the setting without distribution changes. 
LLM based method~(DDI-GPT) still achieves the best performance when distribution changes are introduced.
These findings highlight the importance of accounting for distribution changes in emerging DDI prediction and confirm that the proposed simulation framework is compatible with different drug split strategies.

\clearpage

\subsection{Time and Memory Cost of Evaluated Methods}
\label{app:time_memory}

We provide the time and memory cost of all the evaluated methods in Table~\ref{tab:time_memory}. 
These results are obtained by running the methods on the Drugbank dataset. 
We can see that generally GNN based methods, graph-transformer based method and LLM based method have more time and memory cost than feature based method and embedding based methods.

\begin{table}[H]
    \caption{Time and memory cost of the evaluated methods on Drugbank dataset.}
    \small
    \label{tab:time_memory}
    \begin{center}
        \begin{tabular}{c|cc}
            \toprule
            Method & Time~(min) &  Memory cost~(MB)\\ \midrule
			MLP&6&548\\
			MSTE&16&640\\
			Decagon&23&6736\\
            SSI-DDI & 49 & 3284 \\
            MRCGNN & 836 & 3816 \\
			EmerGNN&725&7431\\
            SAGAN&186&5819\\
            TIGER&293&5734\\
            TextDDI&1871&9404\\
            DDI-GPT&1931&15356\\
            \midrule
        \end{tabular}
    \end{center}
\end{table}

\end{appendices}

\end{document}